%% file: neurips_2020.tex
\documentclass{article}

\usepackage[nonatbib,preprint]{neurips_2021}

\usepackage[utf8]{inputenc} 
\usepackage[T1]{fontenc}    
\usepackage{hyperref}       
\usepackage{url}            
\usepackage{booktabs}       
\usepackage{amsfonts}       
\usepackage{nicefrac}       
\usepackage{microtype}      
\usepackage{xcolor}
\usepackage{verbatim}
\usepackage{caption}
\usepackage{multicol}
\usepackage{multirow}
\usepackage{multibib}
\newcites{app}{Appendix Reference}

\usepackage{amsmath}
\usepackage{graphicx}
\usepackage[ruled,vlined]{algorithm2e}
\usepackage{subfig}
\usepackage{wrapfig}
\usepackage{subfig}

\DeclareMathOperator*{\argmin}{arg\,min}

\newcommand{\loss}{\mathcal{L}}
\newcommand{\data}{\mathfrak{D}}
\newcommand{\E}{\mathbb{E}}

\newcommand{\centerhfill}[1][\quad]{\hspace{\stretch{0.5}}#1\hspace{\stretch{0.5}}}

\newcommand{\jake}[1]{\textcolor{red}{#1}}

\title{A critical look at the current train/test split in machine learning}

\author{
  Jimin Tan \\
  Vilcek Institute of Graduate Biomedical Sciences\\
  New York University\\
  \texttt{tanjimin@nyu.edu} \\
   \And
  Jianan Yang, Sai Wu, Gang Chen, Jake Zhao (Junbo)$^*$ \\  
  Department of Computer Science \\
  Zhejiang University \\
  \texttt{jianan0115, wusai, cg, j.zhao@zju.edu.cn}
  }

\begin{document}

\maketitle
\begin{abstract}
The randomized or cross-validated split of training and testing sets has been adopted as the gold standard of machine learning for decades. The establishment of these split protocols are based on two assumptions: (i)-fixing the dataset to be eternally static so we could evaluate different machine learning algorithms or models; (ii)-there is a complete set of annotated data available to researchers or industrial practitioners.
However, in this article, we intend to take a closer and critical look at the split protocol itself and point out its weakness and limitation, especially for industrial applications.
In many real-world problems, we must acknowledge that there are numerous situations where assumption (ii) does not hold. For instance, for interdisciplinary applications like drug discovery, it often requires real lab experiments to annotate data which poses huge costs in both time and financial considerations. In other words, it can be very difficult or even impossible to satisfy assumption (ii).
In this article, we intend to access this problem and reiterate the paradigm of active learning, and investigate its potential on solving problems under unconventional train/test split protocols. We further propose a new adaptive active learning architecture (AAL) which involves an \emph{adaptation} policy, in comparison with the traditional active learning that only unidirectionally adds data points to the training pool.
We primarily justify our points by extensively investigating an interdisciplinary drug-protein binding problem.
We additionally evaluate AAL on more conventional machine learning benchmarking datasets like CIFAR-10 to demonstrate the generalizability and efficacy of the new framework.

\end{abstract}

\input{sections/intro}

\input{sections/relatedwork}

\input{sections/method}
\input{sections/experiments}

\bibliographystyle{plain}
\bibliography{ref}

\newpage
\input{sections/supplementary}

\bibliographystyleapp{plain}
\bibliographyapp{appendix}

\end{document}

%% file: sections/intro.tex
\vspace{-1em}
\section{Introduction}

Most, if not all, modern machine learning frameworks adopt a fixed and static train/test dataset split. While this split protocol went through massive deployment throughout the years \cite{togersen1999line, leber2017modeling}, we intend to take a more critical look at its limitation and harm in this article. 

Notably, in interdisciplinary fields like AI+drug discovery, most efforts from the machine learning community have been on developing new models or training methods. Almost all the prior publications \cite{ragoza2017protein, gomes2017atomic, ozturk2018deepdta, wang2020deep, wen2017deep, you2019predicting} adopt the classic static setup of train/test split. Hereby we may argue, this has posed some serious systematic errors for the actual drug discovery problem-solving. We list the major discrepancies below:
\begin{enumerate}
\item When starting a drug discovery research process, the domain experts often faces the problem of \textbf{cold-start}, where \textbf{very few} or even \textbf{no} data points are labeled.
\item Unlike computer vision or natural language processing problems, the acquisition of the labels to this problem is significantly harder to many orders of magnitude, because the precise labels must be obtained through lab experiments. In another word, if one insist on positioning the drug discovery problem by the conventional train/test split, it would require the domain experts to conduct laboratory experiments at scale which might not be feasible in most cases. Without any guidance at the early stage, the candidate drug molecules/peptides/proteins have to be chosen randomly from the search space. It is apparent that the lack of guidance could result in unnecessary cost (both time and money). Combined with the large scale of training data required by many models, the random strategy can pose an extremely high cost under a large search space.
\item The conventional static train/test split makes the machine learning side of work to be almost a \textbf{one run} process. This makes little sense in drug discovery due to the domain nature of the subject being highly iterative. The static and fixed split hardly faciliates the machine learning algorithm to interacts with the actual lab experiments conducted by the domain experts.
\end{enumerate}
To this end, in spite of the certain successes of machine learning algorithms proposed in the interdisciplinary fields, we may argue the prior work mostly suffer from a systematic problem that if ignored, often resulting in a reduction of the chances for these techniques to land.

In this article, we make two contributions: 
\begin{enumerate}
 \item We reiterate the less trendy field of active learning, which better bridges with domain experts and better copes with problems that have a \textbf{cold start}.
 \item We propose a novel active learning framework — the adaptive active learning framework (AAL) — that adapts the dynamic data distribution in every run of data exploration. Active learning may suffer from a distribution shift problem which is revealed by our empirical finding in section \ref{section:distribution_shift}, caused by differently distributed sample selections from every run. We implement an instantiation of AAL via incorporating the \emph{deletion} operation besides the \emph{addition} operation in traditional active learning frameworks.
\end{enumerate}

For experiments, we conduct an extensive study on an interdisciplinary protein-drug interaction dataset with our proposed AAL framework. Furthermore, to validate the concept of AAL, we also test it on more traditional computer vision datasets including CIFAR-10. 
It is important to note that we primarily run our experiments using the most vanilla version of the standardized deep or statistical models, and we \textbf{do not tune the hyperparameters} of these models. The main point of this paper is around a new setup for machine learning, taking a data-centered perspective and much less about the model, architecture, or optimization methods. We see our framework being complementary to the advancement of machine learning from other fields, including model evolvement, few-shot learning, semi-supervised learning, etc.

The remainder of the paper is organized as follows: First, we briefly review the related works in the following section. Then we elaborate the concept of AAL and its implementations in Section 3. Experimental results of AAL framework are presented in Section 4. At last, we conclude this paper in Section 5.

%% file: sections/relatedwork.tex
\vspace{-1em}
\section{Related Works}

\subsection{Machine learning in drug discovery}
\vspace{-0.5em}

Machine learning has been applied to problems in many interdisciplinary domains. Drug discovery is one of the main applications in the biochemical domain. Many efforts from the machine learning community have been devoted to improving model performance on tasks like drug-target interaction prediction \cite{ragoza2017protein, gomes2017atomic, ozturk2018deepdta, wang2020deep, wen2017deep, you2019predicting}.  
However, static train/test splits, which are commonly adopted in existing setup from the prior works, are of little use in the real-world deployment: because the training data is almost always accumulated batch-by-batch from a cold start situation,
where the term cold start means very few or even zero available labeled data at the beginning stage of a drug research.
Unfortunately, most of the labeled data sets in this field have already established their own static split and fixed as standardized benchmarks. In spite that the approaches like SimBoost \cite{he2017simboost} and ChemBoost \cite{ozccelik2020chemboost} achieved good performance on KIBA\cite{tang2014making}, BDB\cite{he2016bdb} and other bioactivity data sets, we still found them incompatible with deployments.
Our approach in this paper attempts to break the static split set-up in order to make the computational methods of ML more applicable and deployable.


\subsection{Active Learning}
\vspace{-0.5em}

Despite being a less trendy field, active learning was established as a learning paradigm two decades ago\cite{cohn1994improving}. 
The uncertainty in the predicted distribution is the most popular indicator for data exploration\cite{becker2005two, schohn2000less, scheffer2001active, yang2015multi}.
Query-by-committee is another well-known exploration policy that finds the data point by training a group of different models and compute their predicted differences\cite{seung1992query}. The uncertainty measure for this method can be the Kullback-Leibler divergence \cite{mccallumzy1998employing} or averaged Jensen-Shannon divergence \cite{melville2005active}.
Another method on this line relies on calculating the diversity of a data point in an accumulated training set. This approach aims at diversifying the data points in order to better cover the data manifold possibly with fewer data. There are pre-clustering methods \cite{nguyen2004active, friedman2011active}, link-based methods based on network data \cite{bilgic2009link, bilgic2010active}, matrix-partition \cite{guo2010active}, and bayesian active learning \cite{gal2017deep}, etc.
Although the research in the active learning field has been widely conducted, the primary research direction remains the data exploration policy. The AAL approach we propose in this article focuses less on the exploration, but more on how to \emph{adapt the training set while accumulating}. Since the adaptation procedure is parallel to the exploration strategies, the aforementioned policies are also systematically integrated into the AAL framework.

\vspace{-0.5em}
\subsection{Learning With Insufficient Labeled Data}
\vspace{-0.5em}
There are a considerable amount of efforts being made from the ML community to deal with insufficient labeled data.
For instance, few-shot learning is a popular line of approaches handling the scenario with few data available for each class. Popular approaches may include meta-learning method \cite{finn2017model, jamal2019task},
generative modeling
\cite{zhu2017generative, ducoffe2018adversarial, sinha2019variational},
comparison-based method
\cite{sung2018learning, zhang2018relationnet2},
etc.
Semi-supervised learning has also shown promising results in particular in standard benchmarks like MNIST or CIFAR \cite{rasmus2015semi}. 
This line also includes augmentation-based methods \cite{berthelot2019remixmatch, sohn2020fixmatch}, 
 transductive methods \cite{shi2018transductive, liu2018learning}, 
etc.

In this article, we tend to avoid heavy research or model tuning on these directions. Yet we focus more on the learning paradigm towards real-world ML system deployment, in particular those research challenges involving a cold start scenario. However, these learning methods including few-shot learning, semi-supervised and self-supervised learning can be seamlessly incorporated into our AAL framework. 

%% file: sections/method.tex
\section{Adaptive Active Learning}
\vspace{-0.5em}
\begin{figure}[h]
  \centering
  \includegraphics[width=0.6\linewidth]{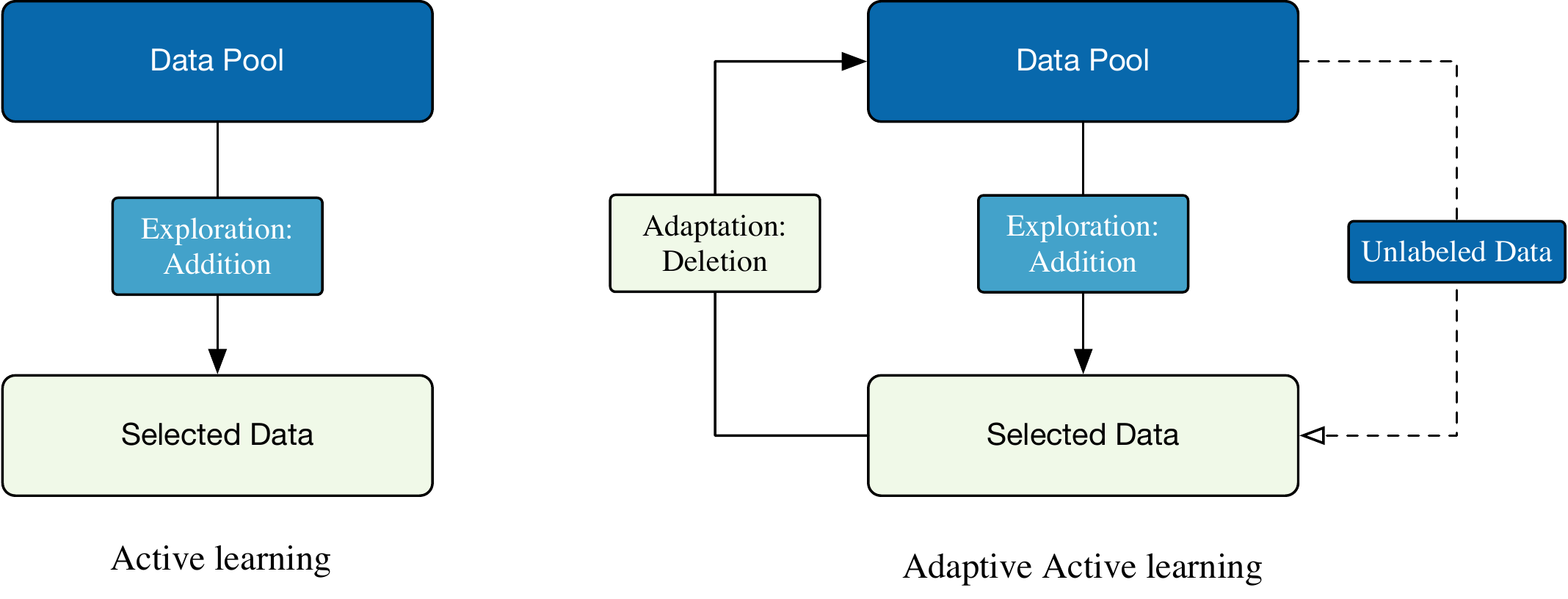}
  \caption{Comparison of active learning and adaptive active learning framework. The unlabeled data in the right plot implies its possible incorporation into the exploration or adaptation policies.}
  \label{fig:aal}
\end{figure}
\vspace{-0.5em}

Many tasks in the industry and especially interdisciplinary applications do not have data readily available. The conventional static train/test split required by most machine learning algorithms does not suit the need.
The active learning framework \cite{cohn1994improving} facilitates synchronization between data-collection and model-training processes, and iteratively improves both over time.
Albeit its superiority over a random data selection policy, the active learning framework still has some drawbacks. For instance, when dealing with a data cold-start scenario at the beginning stage of a drug-discovery research program where the domain experts are presented with zero or very few available data points, the initial exploration policy could perform poorly or have an unstable estimation.
\emph{Unidirectional} addition operation prone to a drastic distribution shift problem where the new batch of data differs largely from the previous training set. Yet due to the monotonic and additive nature of active learning, it is not obvious how the distribution shift could be corrected in the vanilla AL framework.

Therefore, we propose an enhanced system: Adaptive Active Learning, dubbed as AAL. Parallel to the addition operation, by augmenting the conventional framework with an \emph{adaptation policy}. There numerous ways to implement the adaptation policy, but we instantiate it with the simplest form — deletion. The deletion mechanism removes ill-behaved data or outliers from the training pool right after adding a new batch of data points. 
We leave the other adaptation policies to future exploration. 

\vspace{-0.5em}
\subsection{An AAL Instantiation with deletion: AAL-delete}
\vspace{-0.5em}

In this work, to present the meaningful incorporation of the adaptation policy. 
As we mentioned, we choose the simplest form of it by allowing it to delete data points from the training pool.
As a result, AAL-delete is an iterative procedure that performs data points addition and deletion alternatively at each round.
A pseudo code is displayed in Algorithm~\ref{add_with_delete}.


\begin{algorithm}[H]
\small
\label{add_with_delete}
\SetAlgoLined
\SetKwInOut{Require}{Require}
\SetKwInOut{Ensure}{Ensure}
\Require{Unlabeled data pool $\mathcal{U}$, labeled data pool $\mathcal{L}$, number of steps $T$, start size $M_0 \ll |\mathcal{U}|$, number of samples added per cycle $N_a$, number of samples deleted per cycle $N_d$, Scoring policy for addition $\mathcal{S}_a$, Scoring policy for deletion $\mathcal{S}_d$, continue criteria $\mathcal{O}_T$.}
\While{$O_t$}{
$S_a$ = $\mathcal{S}_a(x_u)$, $\forall x_u \in \mathcal{U}$; $S_a$ is defined as addition scores

$\mathcal{B}_a$ = $\{x_u | x_u \in $ top $N_a$ elements ranked by $S_a$ in $\mathcal{U} \}$; Query top rated samples

$\mathcal{L}=\mathcal{L}\cup\mathcal{B}_a,\mathcal{U}=\mathcal{U}\backslash\mathcal{B}_a$;

$\mathcal{B}_d$ = $\{x_l | x_l \in $ last $N_d$ elements ranked by $S_d$ in $\mathcal{L} \}$; Query ill-behaved samples

$\mathcal{L}=\mathcal{L}\backslash\mathcal{B}_d,\mathcal{U}=\mathcal{U}\cup\mathcal{B}_d$;

$O_{t+1} = \mathcal{O}_t(\mathcal{L})$
}
 \caption{AAL-delete framework}
\end{algorithm}
\vspace{-0.5em}



\subsection{Data addition and deletion policies}
\vspace{-0.5em}

Defining the quality of each data point with various metrics is crucial for AAL to select the optimal batch in the unlabeled data pool to add or to delete from the current pool. More so, the policies determine the model performance which impacts the next round of iteration in AAL. 
In theory, the policy used for adding and deleting can be different depending on the task, and can be any functional forms.


\textbf{Entropy.} This is defined as Shannon entropy,
 which entails a confusion state between different classes. For example, in a cat/dog image classification problem, an image of an animal that has features of both dog and cat could yield high entropy. In our experiments, we demonstrate that identifying and learning confusion data points can enforce the classifier to be more accurate and generalizable. But entropy can be difficult to quantify in a continuous space.

\textbf{Feature diversity.} Aside from data points that are mixtures of different classes, data points that are generally far away from the center of the data manifold also provide extra information. Previous works have demonstrated that selecting subsets from clustering data by the distance between their learned representations can result in a smaller dataset with similar representation power \cite{birodkar2019semantic}. For dealing with high-dimensional feature vectors, we choose an angle-based metric called cosine distance:
$$
Dist(x) = 1 - \frac{
	\left \langle 
	f(x, \theta), \overline{f}(\mathcal{X}, \theta)
	\right \rangle
	}
{\Vert f(x, \theta)\Vert_2 \Vert \overline{f}(\mathcal{X}, \theta) \Vert_2},
$$
where $\overline{f}(\mathcal{X}, \theta)$ denotes the average feature vector, 
$\mathcal{X}$ is the set of all training samples, $f(\cdot,\theta)$ is the backbone feature extractor. 
In AAL, we expect this policy to select data points away from the current center of the training pool (in representation space) in order to faciliate distribution coverage with higher data efficiency.

\textbf{Uncertainty.} The last criterion concerns the uncertainty of a data point. The corresponding query strategy is commonly known as query-by-committee(QBC)\cite{seung1992query}. Training multiple models at each round of AAL-delete can be computationally challenging, especially for the deep neural network models. It has been shown that Dropout can represent uncertainty in deep learning models \cite{gal2016dropout}. Thus, we rely on the Dropout layer and turn the eval mode off during the inference process for the specific implementation. 
For a classification problem, given a committee with $C$ members $\mathcal{C} = \{\theta^{(1)},\dots,\theta^{(C)}\}$ in evaluation, the uncertainty metric is defined as follows:

\vspace{-2em}
\begin{align*}
JS(p_1,\dots, p_C, x) & = \frac{1}{C}\sum_{c=1}^{C}KL(p_c(x)||\overline{p}(x)), 
\end{align*}
\vspace{-1em}

where $p_c$ denotes $P(\cdot|\theta_c)$, the output distribution of the $c^{th}$ model in our committee, when $\overline{p}(x)$ represents their average probability distribution. We present $KL$ to indicate KL-divergence, when $P(y|x)$, $Q(y|x)$ denote the posterior probability of each label $y$ when given data point $x$, corresponding to two given distributions $p(x)$, $q(x)$ respectively. When all distributions given by models in the committee are exactly the same, JSD is at its minimum value zero. Likewise, for regression problem, we yield this metric simple by calculating the variances of all the outputs from the model committee.

\textbf{Ensemble}
In AAL-delete, it is crucial to balance these different metrics. 
Since these metrics are generally not normalized or upper-bounded, we primarily adopt a ranking mechanism. Thus a hybrid framework combining variaties of existing strategies has been made to assist AAL-delete.

%% file: sections/experiments.tex
\section{Experiments}

We used two benchmarking datasets with different tasks respectively. We start from KIBA\cite{tang2014making}, a regression task on protein-drug binding affinity. We then extend the evaluation to image classification tasks including CIFAR-10. 

\vspace{-1em}
\subsection{Protein-drug affinity discovery}
\label{section:affinity}
\vspace{-0.5em}

The KIBA dataset is named after the KIBA method which produces a score from different kinase inhibitor bioactivity sources including $K_i$, $K_d$ and $IC_{50}$. We adopted the preprocessing pipeline from a previous study\cite{ozturk2018deepdta}. The preprocessed KIBA dataset includes 229 proteins and 2111 drugs and in total 118254 measured KIBA score between proteins and drugs. Both drug and protein are encoded as one-hot vectors. 

\textbf{Metrics.} Our aim is to cover desired data with fewer data assessed than randomly selecting data points. In KIBA, the coverage score $C_{1k}$ is defined as follows:  $\bar{\mathcal{D}}_{1k}$ is the top 1000 prediction pair and $\mathcal{D}_{1k}$ is the top 1000 ground truth, the score is the ratio between correctly selected protein-drug pair and actual protein drug pair.
\[C_{1k} = \frac{|\bar{\mathcal{D}}_{1k} \cup \mathcal{D}_{1k}|}{|\mathcal{D}_{1k}|}\]

\begin{wrapfigure}{r}{0.5\textwidth}
\centering
\includegraphics[width=1\linewidth]{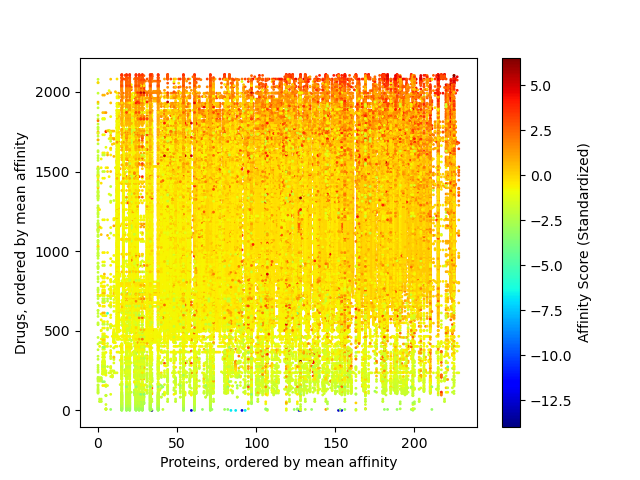}
\caption{Heatmap of affinity scores, protein and drug axis are ranked by mean affinity scores.}
\label{figure:affinity_score}
\end{wrapfigure}

\textbf{Active learning framework.} We adopt an active learning framework to approach the drug discovery problem. We initially select 64 drug-protein pairs and reveal the corresponding affinity scores. A model is trained on the initial data and later makes inferences on the full dataset to obtain all the metrics we need for the next data batch selection. In KIBA, we used uncertainty metrics to determine which data points are added or deleted from the current data batch. After computing the metrics on the full dataset, 64 data is selected from the unlabeled pool to join the current labeled data. For deletion policy, the bottom 8 data points with the lowest scores are removed from the labeled data in each iteration.
In this paper, we only focus on the data selection process in AAL, and we do not explore details like model architectures or training techniques. For this task, we implemented a model that requires minimal computation power as our experiments require repeated training over multiple iterations. More experiment details are included in supplementary materials. 

\textbf{Random and greedy baselines.} We compare our method with two baseline exploration strategies: random and greedy. In the random baseline, we select data from the unlabeled pool for forming the next iteration batch in an unbiased way. The greedy strategy uses the current model to evaluate all protein-drug binding and select the highest scoring pairs. In this study, an increment of 64 samples is implemented.

\begin{wraptable}{r}{0.5\textwidth}
\centering
\small
\begin{tabular}{ ccc } 
\toprule
Exploration Strategy & Data Quantity \\
\midrule
Random & > 19136 \\ 
AL Greedy & > 19136 \\ 
AL Hybrid & 10816 \\ 
\textbf{AAL Hybrid} & \textbf{9408} \\
\bottomrule
\end{tabular}
\caption{Data quantity needed to reach coverage score of 0.3.}
\label{table1}
\vspace{-1em}
\end{wraptable}

\textbf{Hybrid strategy with uncertainty.} 
To select the best protein-drug pair with high-affinity scores, we argue that two criteria must be met. First, we need to look deeper into specific drugs or proteins that generally outperform others in terms of affinity. However, given that we are starting from scratch, it is likely that the model is trapped at some local minimum. Thus, we also need to explore as many other drug-protein options as possible. This strategy shares the core concept with the exploration/exploitation balance in reinforcement learning.
We adopted a half split in the hybrid strategy. 32 out of 64 data points are selected greedily based on their predicted affinity scores. The rest are selected based on how \textit{uncertain} the model perceives data points. The uncertainty metrics,  can be defined as a standard deviation of prediction in regression problems. We trained five models with random initialization in each iteration. The uncertainty is measured by the variance between five different model predictions.

\textbf{AAL with deletion.} The traditional active learning method only considers adding data to the labeled set. Considering the stochastic nature of the active learning process, data points that are less optimal for improving model performance might be added. Thus, we adopted AAL with deletion to remove non-optimal datapoints and ensure that the labeled data pool stays coherent and efficient. 

\subsubsection{Convergence on high affinity drug-protein pair}
\vspace{-0.5em}

\begin{figure}
  \centering
  \includegraphics[width=0.6\linewidth]{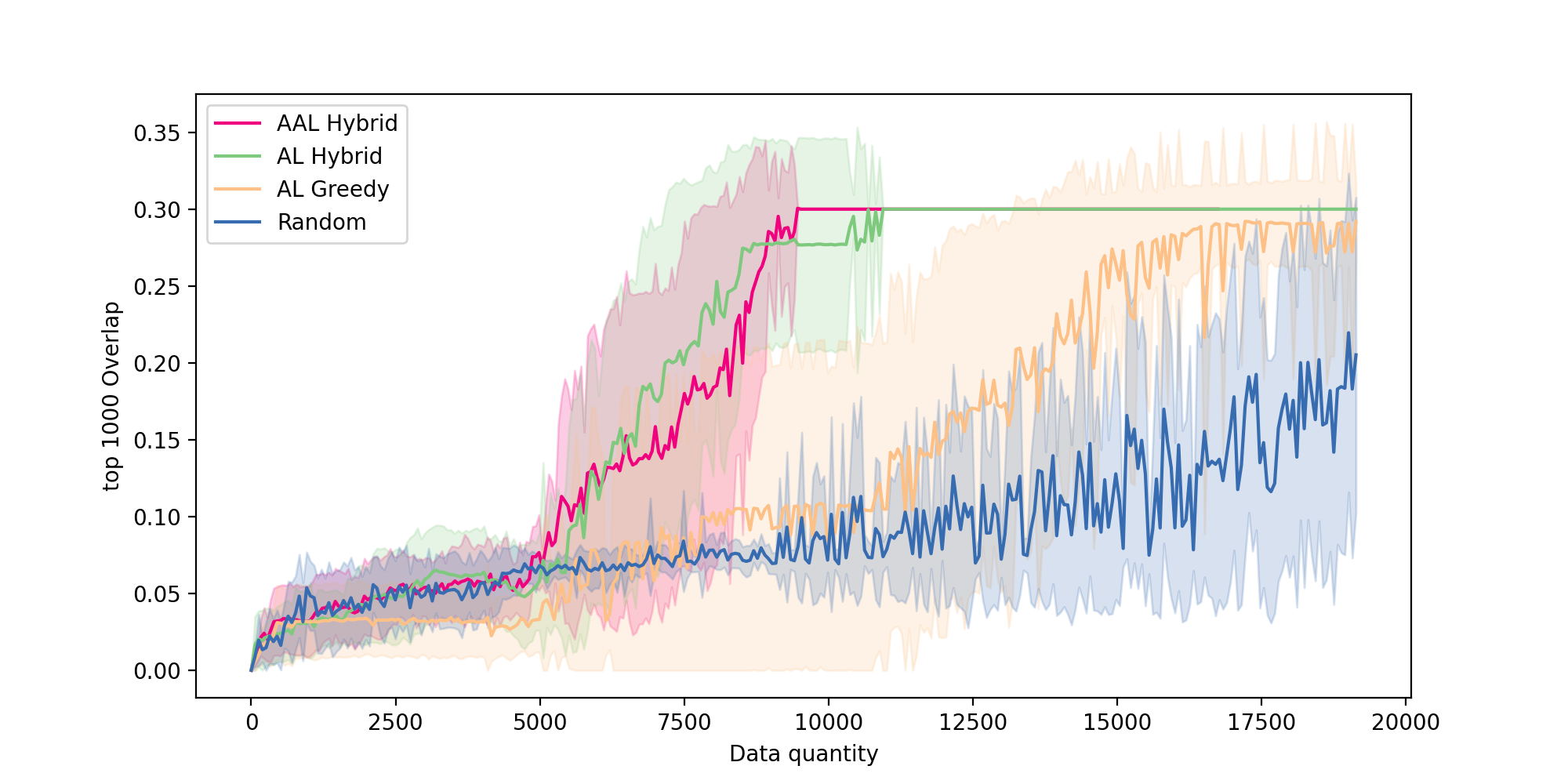}
  \caption{Mean coverage score of four different strategies shown in the legend. Shaded regions are error of 1 std of the corresponding mean strategy scores.  Each strategy was averaged from 10 experiments.  Experiment stops when coverage score reaches 0.3. AAL Hybrid outperforms AL Hybrid and reaches the target coverage score of 0.3 faster. The greedy and random baselines are much slower.}
\label{figure:1}
\end{figure}

\begin{table}
\centering
\small
\begin{tabular}{ ccccc } 
\toprule
Strategy & \multicolumn{4}{c}{Coverage Score ($C_{1k}$)} \\
\cmidrule{2-5}
 & $s = 3000$ & $s = 6000$ & $s = 9000$ & $s = 12000$ \\
\midrule
Random & 
$0.049 \pm 0.019$ &
$0.065 \pm 0.012$ &
$0.070 \pm 0.007$ &
$0.107 \pm 0.066$ \\ 

AL Greedy & 
$0.032 \pm 0.022$ &
$0.071 \pm 0.077$ &
$0.092 \pm 0.106$ &
$0.151 \pm 0.129$ \\ 

AL Hybrid & 
$\mathbf{0.059 \pm 0.030}$ &
$0.111 \pm 0.067$ &
$0.277 \pm 0.070$ &
$\mathbf{>0.300^*}$ \\ 

\textbf{AAL Hybrid} & 
$0.054 \pm 0.019$ &
$\mathbf{0.120 \pm 0.085}$ &
$\mathbf{0.284 \pm 0.047}$ &
$\mathbf{>0.300^*}$\\  
\bottomrule
\end{tabular}
\caption{Coverage score at four different data quantity level from 3k to 12k. AAL Hybrid outperforms all other methods. AL Hybrid outperforms AL Greedy and Randome by a large margin. *Iteration stops at coverage score of 0.3, and we assume extra data does not compromise model performance.}
\label{table2}
\vspace{-2.5em}
\end{table}

\textbf{Reduced data requirement.} AAL with the hybrid strategy gives us the best performance overall. (Fig. \ref{figure:1}) Active learning with a hybrid strategy outperforms both baselines. In baseline, greedy outperforms random baseline after the labeled set includes more than 5000 data points. Fixing the coverage score at 30\%, AAL hybrid only utilizes less than 50\% of the data compared with both greedy and random baseline.(Table \ref{table1}). With AAL we can obtain the theoretical target coverage less than a third of the time comparing to greedy or random baseline.

\textbf{Higher coverage at a given data quantity.} We looked at the performance at different data quantities.(Table \ref{table2}) The hybrid model with deletion consistently outperforms the two baseline methods by a large margin, covering more than 3 times as many top affinity pairs in comparison. Hybrid with deletion also outperforms hybrid strategy without deletion consistently at $s = 9000$.

\vspace{-0.5em}
\subsubsection{Exploration trajectory}
\label{section:exploration}
\vspace{-0.5em}

We investigated the large performance gap between different strategies. Specifically, we explore the difference between AAL and active learning with a greedy-only policy.

\textbf{Ranked affinity grid.} To visualize the trajectory of the learning process, we need to define a 2D grid to place  all the data points.
Then, we rank drugs and proteins based on their average affinity score of all possible combinations in the database. The result is a grid with higher affinity pair on the top right and lower score pair on the bottom left. We show a heatmap of this constructed grid colored by standardized affinity scores. (Figure \ref{figure:affinity_score})

\begin{figure}
\centering
\includegraphics[width=0.7\linewidth]{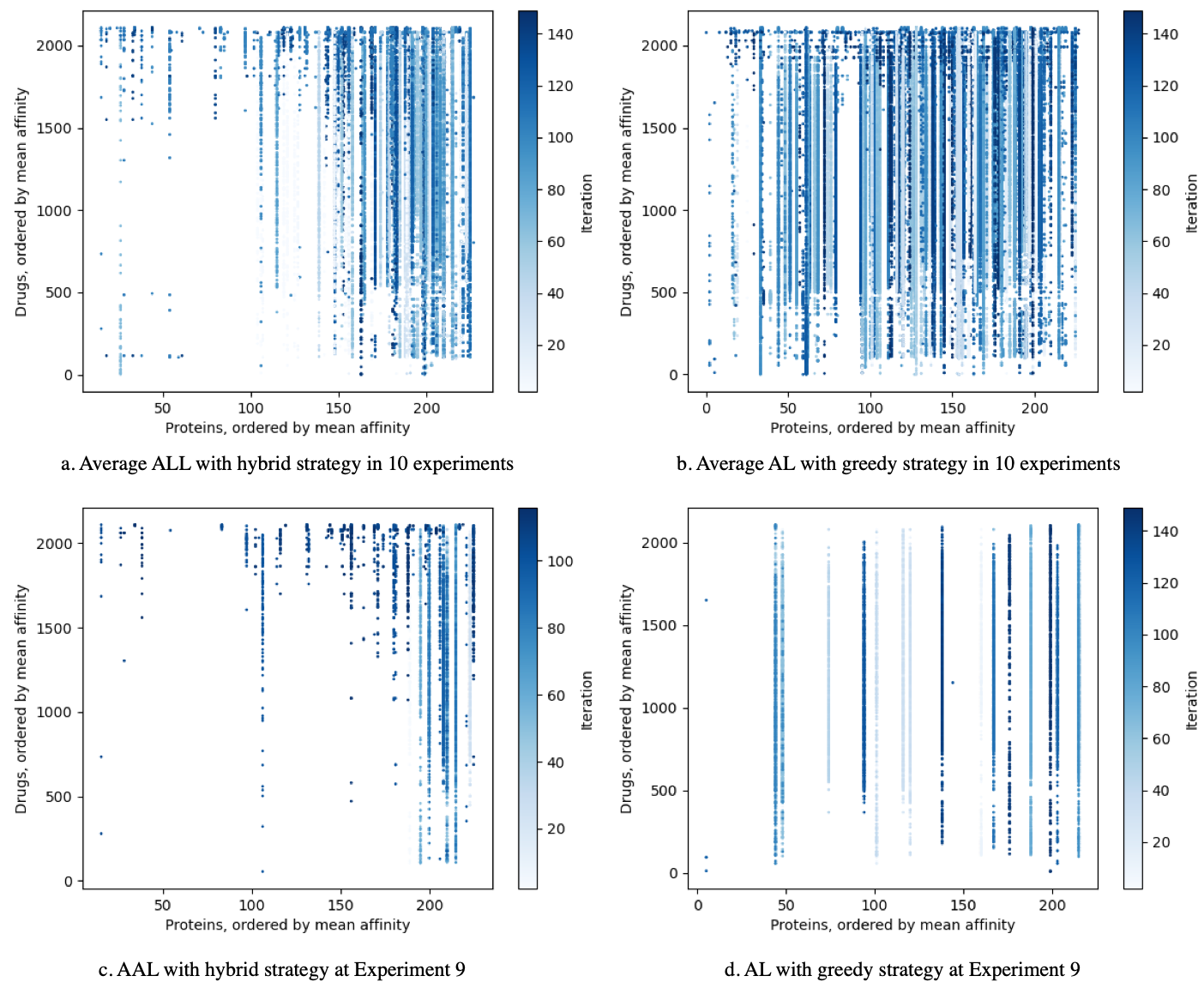}
\caption{Exploration trajectories of AL and AAL. See section \ref{section:exploration} for more details}
\label{fig:kiba_explore}
\end{figure}

\begin{wrapfigure}{r}{0.5\textwidth}
    \centering
    \includegraphics[width=1\linewidth]{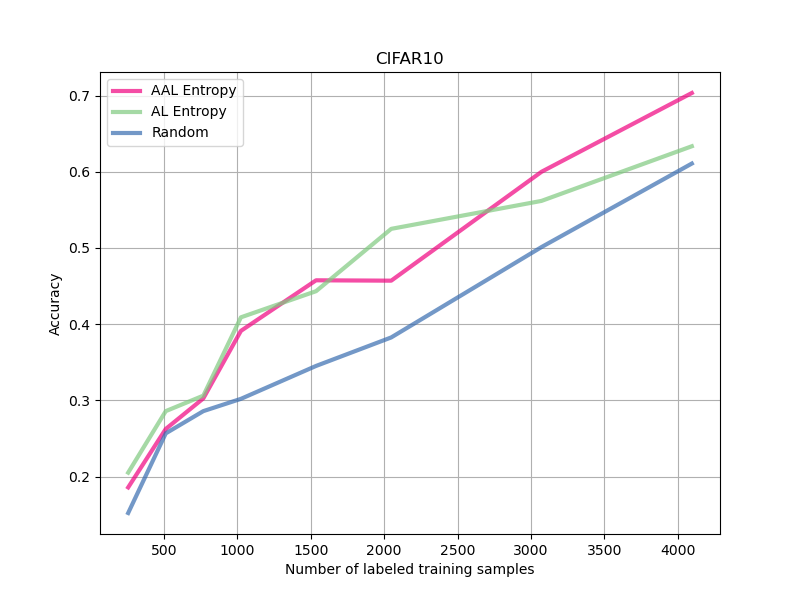}
    \caption{Comparsion of different sampling methods trained with ResNet-18 on CIFAR-10.}
    \label{cifar_accu}
\end{wrapfigure}

\textbf{Greedy, but in a different way.} The greedy strategy selects the data with the highest affinity scores. In our hybrid strategy, half of the data are also selected using a greedy approach. We compared the data selected by the greedy-only strategy with the hybrid strategy. The latter is doubled in quantity to match the data selected by the greedy-only method. We find that on average, the greedy-only strategy covered the whole data manifold. (Fig. \ref{fig:kiba_explore}b) In comparison, the hybrid strategy focuses on the top right region where the highest affinity pair lies (Fig. \ref{fig:kiba_explore}a). An inspection of a single experiment yields more information about the failure of the greedy-only method. Notice that the greedy trajectory stays in one protein until it exhausts most combinations with other drugs. (Fig. \ref{fig:kiba_explore}d). The hybrid strategy explores more types of proteins but less on a single protein except for high-affinity proteins. (Fig. \ref{fig:kiba_explore}c). The difference in trajectory confirmed our hypothesis that the greedy-only method is likely to get trapped in local minimum while the hybrid approach is more balanced in exploration and exploitation, achieving superior results. More single experiment data are provided in supplementary materials.

\subsection{Image classification}

To verify that AAL can generalize outside of drug discovery, we evaluate its performance on the image classification tasks. We choose a widely used benchmarking computer vision dataset, CIFAR-10, 

\textbf{Baseline policies.} We choose two baseline methods on top of AAL for comparison: random selection, and traditional active learning. We use a combination of entropy-based uncertainty and distance-based uncertainty.  We adopted different strategies for adding and deleting to maximize performance. Entropy is used as the sole metric for adding data, while deletion is determined by a combination of entropy and distance score. More detailed experiment setup is available in the supplementary material.


\textbf{Results.} The empirical results are demonstrated in Fig \ref{cifar_accu}. 
Both AL and AAL outperform the random baseline. AAL and AL are close in the beginning and AAL performs better as trainig set grows.


\subsubsection{Exploration Trajectory}

\begin{wrapfigure}{r}{0.4\textwidth}
    \centering
    \includegraphics[width=1\linewidth]{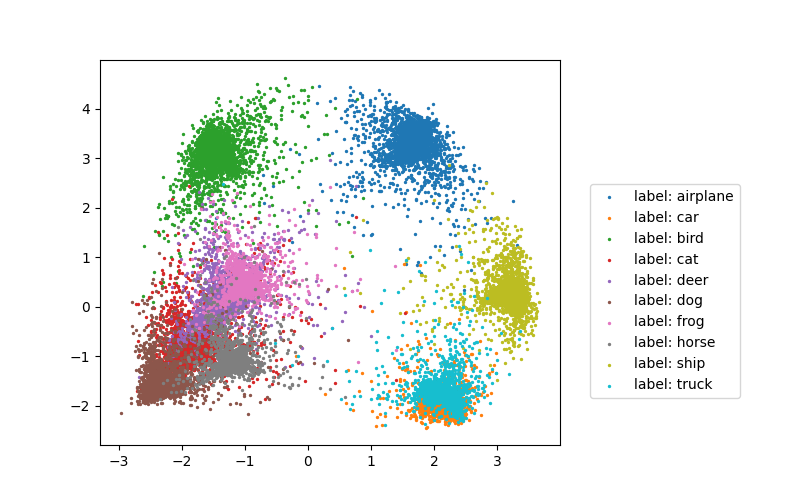}
    \caption{2D PCA on the feature vector of each data point.}
    \label{fig:cifar10_class_distributions_pca}
\end{wrapfigure}

\begin{figure}
    \centering
    \subfloat[CIFAR10, t-SNE]{
        \includegraphics[width=0.22\linewidth]{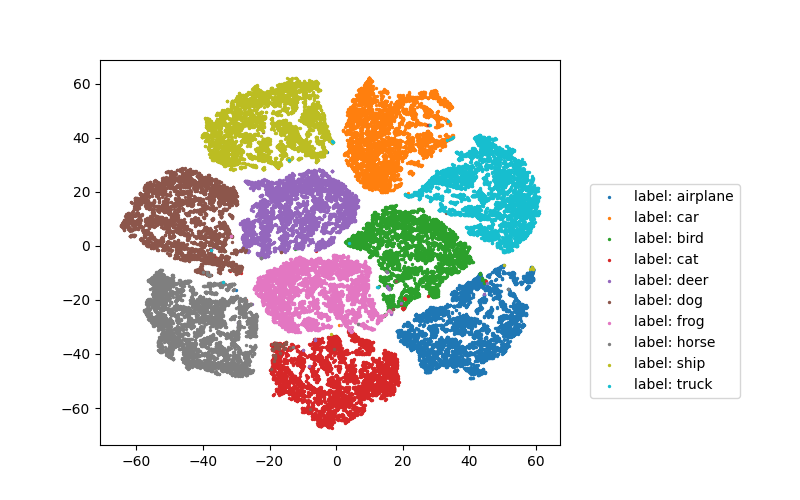}
    }
    \subfloat[AAL at iter 20]{
        \includegraphics[width=0.22\linewidth]{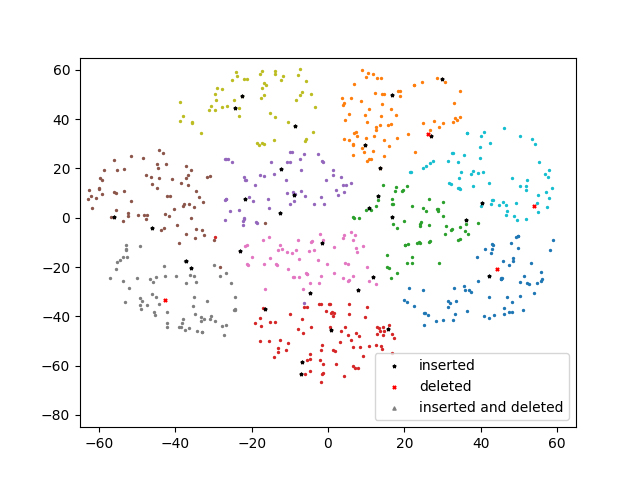}
    }
    \subfloat[Data added]{
        \includegraphics[width=0.22\linewidth]{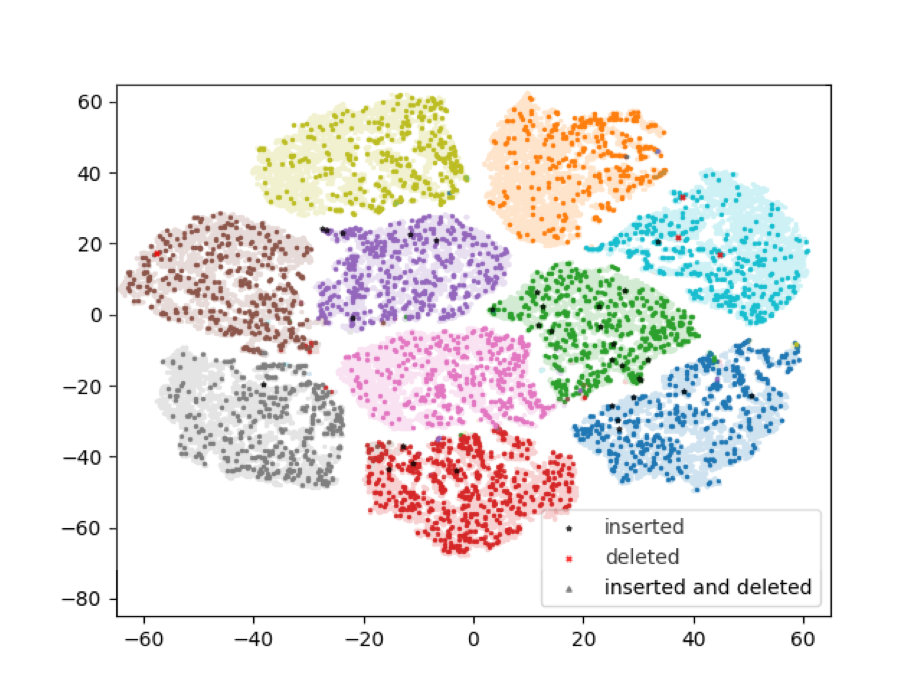}
    }
    \quad
    \subfloat[Data deleted]{
        \includegraphics[width=0.22\linewidth]{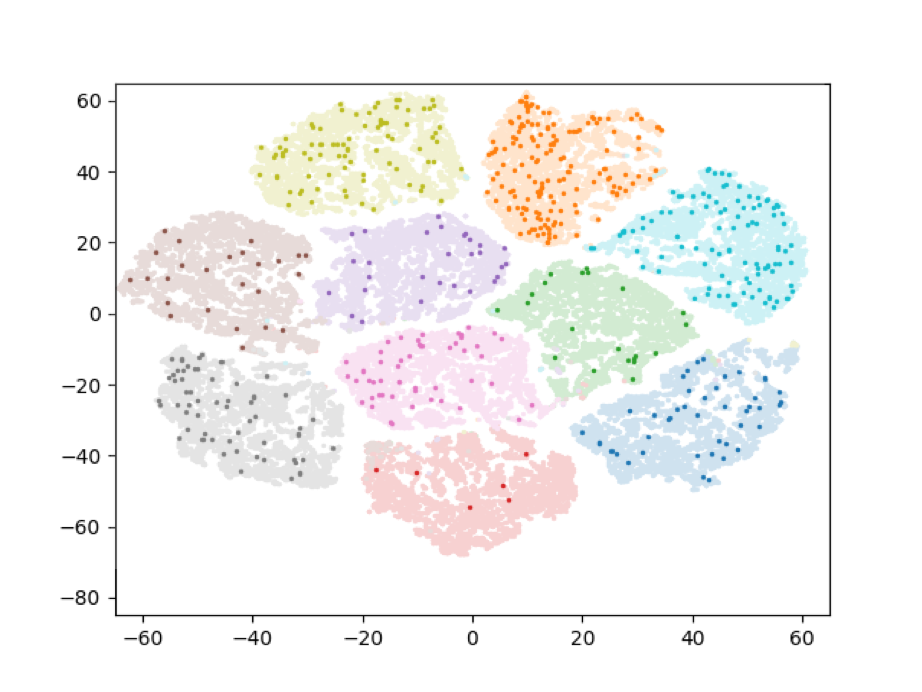}
    }
    \caption{Exploration trajectory for a single experiment in CIFAR-10.\vspace{-1em}}
    \label{fig:cifar_last}
\end{figure}

We focus on CIFAR-10 as a common dataset for evaluation.
To understand what data points are chosen at a given iteration, we constructed a 2D space where all the data can be placed accordingly. Different from the KIBA dataset, where each protein and drug is represented as a one-hot vector, image data can be represented as meaningful embedding vectors by processing them with a pre-trained convolutional feature extractor. We extracted the embeddings using a  ResNet-18 feature extractor, and project the resulting vectors onto a 2D space with t-SNE using a perplexity of 30.0. The result is a 10 cluster mapping of all the images.(Fig. \ref{fig:cifar_last})

\textbf{Adding confusing samples.} In CIFAR-10, we use entropy as the main criterion for selecting new data. As mentioned above, entropy measures the level of confusion between different classes which is an inherent property of the datapoint. Thus, hypothetically the selected data with this strategy lies between decision boundaries. t-SNE visualization does not proportionally represent distance between clusters, but within the cluster similar points on high dimensions are closer. We expect data selected by entropy will be further away from each centroid. Based on the distribution of selected data points at 20th iteration (Fig. \ref{fig:cifar_last}), most class is less dense in the center comparing to the t-SNE of the full dataset.

\begin{figure}
  \centering
  \includegraphics[width=1\linewidth]{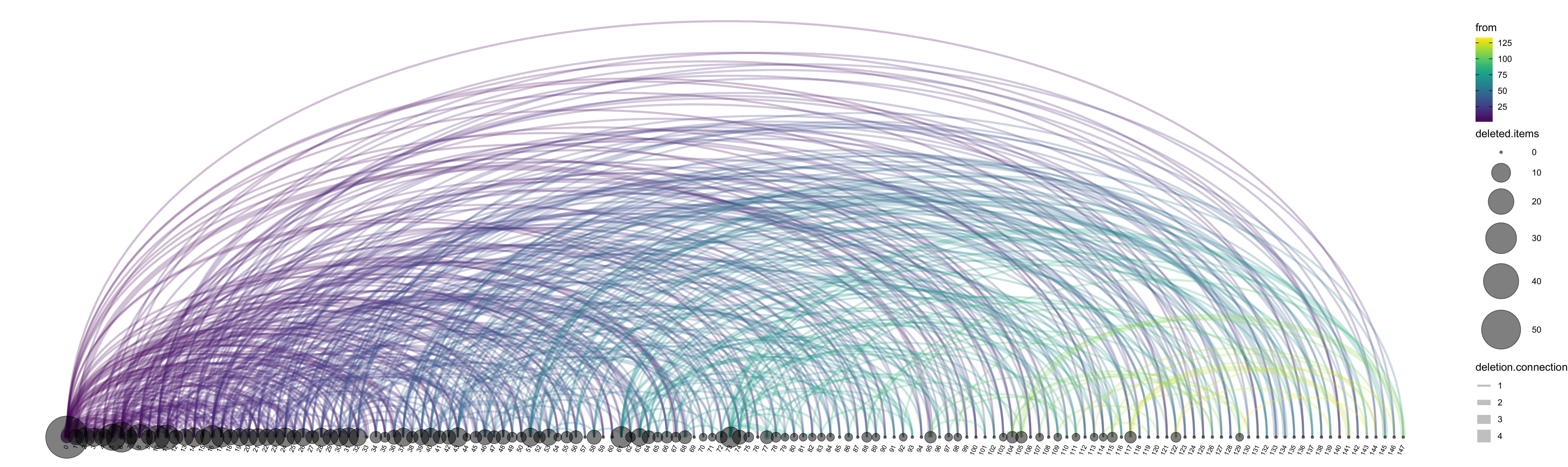}
  \caption{A network of deleted data point origins. A node at position $i$ represents an iteration $t_i$. Each edge is directed from right to left side connecting two nodes. A connection from $t_m$ to $t_n$ with thickness $s$ represent that $s$ numbers of data deleted at $t_m$ are originally added at $t_n$. The size of $t_i$ represent the total number of data that will be deleted in the future and added at step $t_i$. \textbf{Zoom-in is recommended.}}
  \label{cifar10_single}
\end{figure}

\textbf{Deletion of common items.} The metrics used for deletion are combinations of both entropy and cosine distance, and data that are common or typical among the class will be the target for deletion. We expect the deleted items to be closer to the centroids and complements added items. In (Fig. \ref{fig:cifar_last}) The deleted data are more evenly distributed and closer to the center of each cluster compared to added data which tends to gather around boundaries.  
In CIFAR-10, we did not adopt a hybrid training approach since the goal is to improve the overall accuracy, and therefore no exploitation is needed. The difference in metrics used for deletion is reflected on deletion patterns across iterations. In image classification, images added at earlier iterations are deleted more often and as the learning continues, data added in later iterations have less chance of removal. (Fig. \ref{cifar10_single}) The model trained with few data in early iterations is less capable and the metrics which are derived from the model are less accuracy. As training data increases in size, the model learns a better representation, and data of greater quality are selected. As a result, poor data selected at the beginning are discarded.

\begin{figure}
    \centering
    \subfloat[AL, $s=128$]{
        \includegraphics[width=0.18\linewidth]{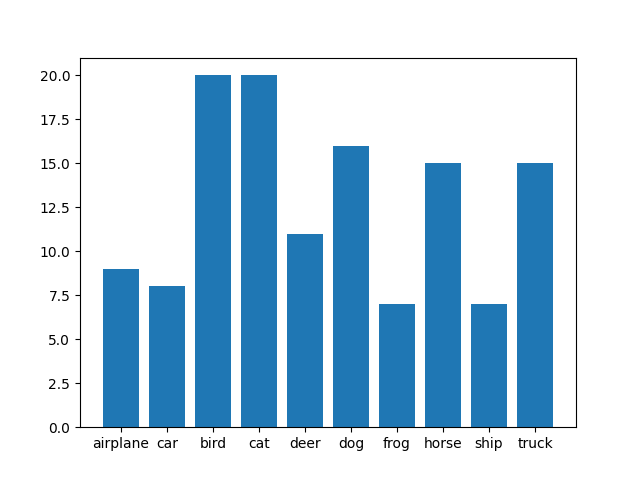}
    }
    \subfloat[AL, $s=268$]{
        \includegraphics[width=0.18\linewidth]{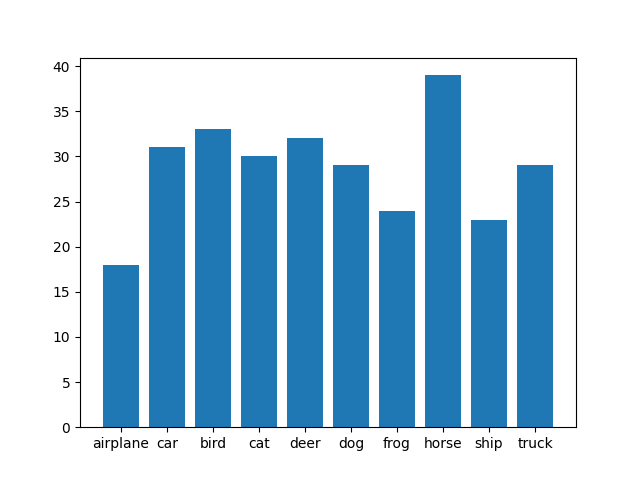}
    }
    \subfloat[AL, $s=1052$]{
        \includegraphics[width=0.18\linewidth]{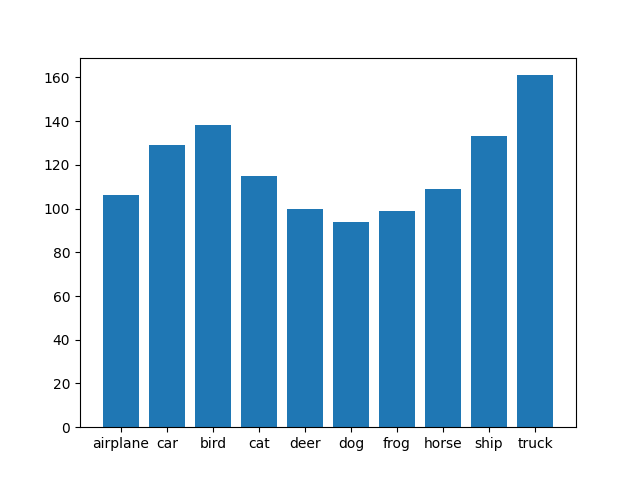}
    }
    \subfloat[AL, $s=4104$]{
        \includegraphics[width=0.18\linewidth]{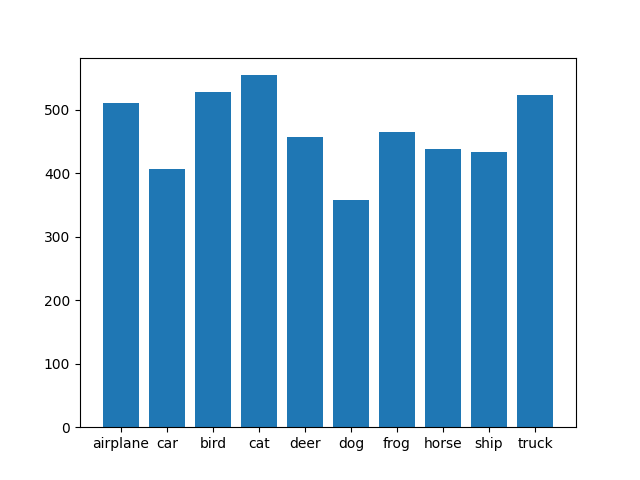}
    }
    \subfloat[AL, $s=8192$]{
        \includegraphics[width=0.18\linewidth]{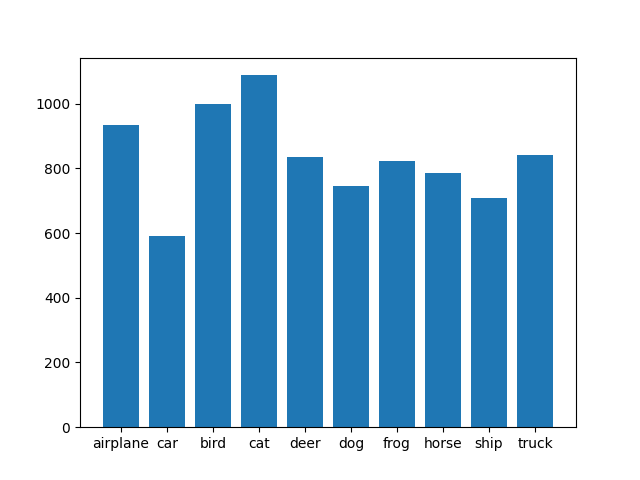}
    } \newline
    \subfloat[AAL, $s=128$]{
        \includegraphics[width=0.18\linewidth]{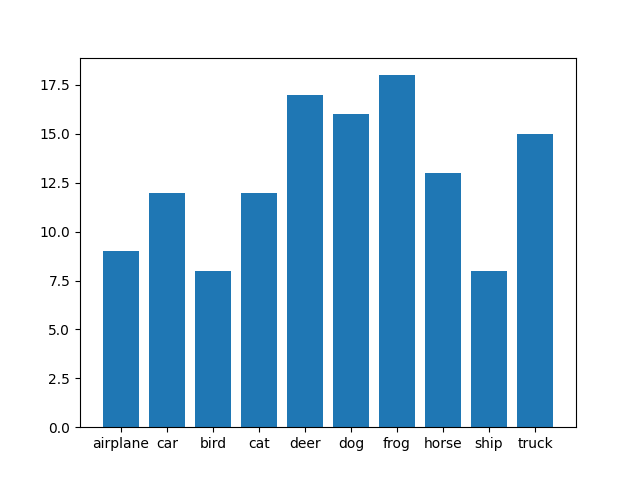}
    }
    \subfloat[AAL, $s=268$]{
        \includegraphics[width=0.18\linewidth]{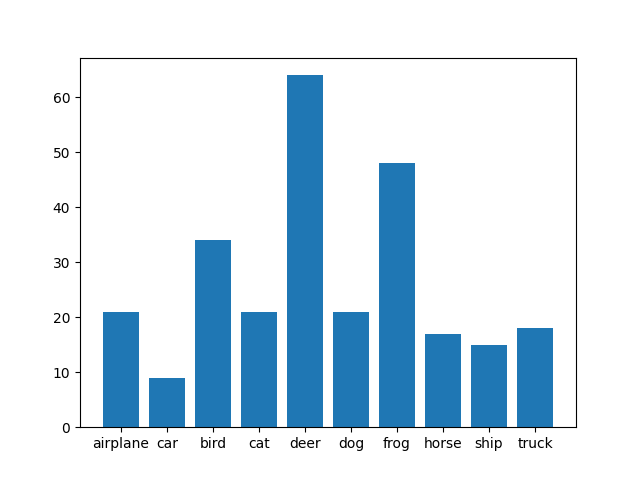}
    }
    \subfloat[AAL, $s=1052$]{
        \includegraphics[width=0.18\linewidth]{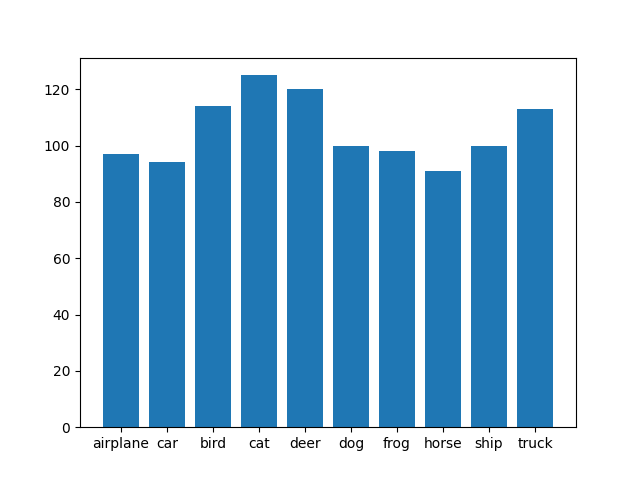}
    }
    \subfloat[AAL, $s=4104$]{
        \includegraphics[width=0.18\linewidth]{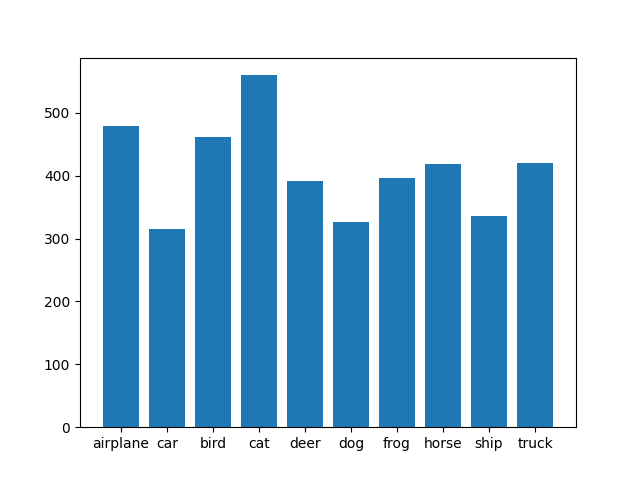}
    }
    \subfloat[AAL, $s=8192$]{
        \includegraphics[width=0.18\linewidth]{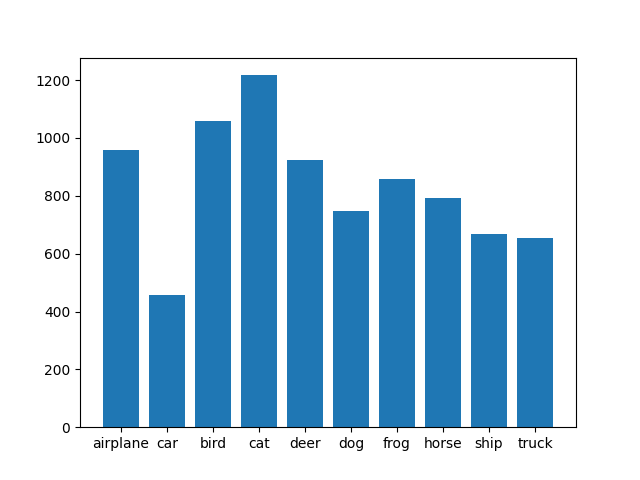}
    }
    \caption{This group of image shows how data distribution changes as AL and AAL iterate, from a randomized initial pool of 128 samples ($s=128$) to 8192 samples.\vspace{-1em}}
    \label{fig:cifar10_class_distribution_changes}
\end{figure}

\subsubsection{Data distribution shift}
\label{section:distribution_shift}

In this section, we describe how data distribution shifts in CIFAR-10 during the data selection process of AL and AAL. Data distribution for each method is visualized by a barchart of different classes selected. 

We examined the class balance of AL and AAL at four different iterations. Both strategies experience major changes before settling down on the final distribution. Notice that in the last iteration, both AL and AAL stabilize to similar distributions with the \texttt{Cat} category being the most selected class. (Fig. \ref{fig:cifar10_class_distribution_changes}) According the trajectory, both strategies favor cats because they resembles dogs, frogs and horses according to PCA in the feature space (Fig. \ref{fig:cifar10_class_distributions_pca}), making them the main targets for our entropy based methods. In the deletion t-SNE clustering, we also see very few cat images removed from selected data (Fig. \ref{fig:cifar_last}b).

The distribution shift is much more drastic at earlier stages (Fig. \ref{fig:cifar10_class_distribution_changes}b, c and Fig. \ref{fig:cifar10_class_distribution_changes}g, h). In adjusting distributions, AAL behaves more aggressively than AL. The most selected category, \texttt{Cat} is favored by AAL early on. For AL, the most selected category shifts gradually from horse to truck and to cat in the end. Comparing the final distribution selected by both strategies, more distinctions are made between different classes in AAL than AL. Without any adaptation policy, vanilla AL accumulates all data points, resulting in a lower signal to noise ratio.

%% file: sections/supplementary.tex
\section{Appendix}

\subsection{Important Notices on the Experiment Protocols}

\subsubsection{On the metric for KIBA}

As stated in section \ref{section:affinity}, we developed a metric measuring the top affinity-scored candidates among the ground-truth. 
We found several domain papers to support our metric development~\citeapp{di2015strategic, eustache2016progress, miranda2008identification}, and we acknowledge the discrepancy from the default choice in the ML community.

\textbf{Why not RMSE?} While the KIBA dataset is often treated as a common regression task from ML cohorts~\cite{ragoza2017protein, gomes2017atomic, ozturk2018deepdta, wang2020deep, wen2017deep, you2019predicting} , we argue this is very far from the true goal of the drug discovery. Simply put, in drug discovery, the actual deployment \textbf{cares little} about how accurate the affinity score is being predicted. Nevertheless, it emphasizes the laboratory performance of the selected drug candidates based on the predicted scores. That is, rather than using RMSE or its variants which could be overly general, we devise and rely on a ranking version of the coverage score.

\subsubsection{On the distribution shift problem}
\label{section:app_shift}
To complete Figure~\ref{fig:cifar10_class_distribution_changes} with some more quantitative results,  \label{distribution_shift} Table \ref{CIFAR_10_KL} shows the mean KL divergence of the dynamic training set distributions between different checkpoints. 
Namely, assuming a learning process contains 100 iterations, the result corresponding to 0\%-10\% denotes the KL divergence between the training set distributions at the 10th iteration and the training set distribution of the very initial training pool.

Note that because of the insurmountable difficulty to estimate the KL divergence between two high-dimensional datasets, we reduce it to use their corresponding data label distributions. Similar techniques are also seen in image generation quality estimation~\citeapp{NIPS2016_8a3363ab}.

This result notably demonstrates that the AAL explores more aggressively in the beginning stage than AL, while converging the distribution much quicker towards the end (lower KL between 10\% and 100\%). Therefore, we may conclude that the AAL paradigm can perform more robustly against the distribution shift problem. Note that the result in Table 3 coincides with our observations in section~\ref{section:distribution_shift}. 





\begin{table}[h]
	\centering
	\begin{tabular}{cccccc}
        \toprule
		Algorithms & \multicolumn{2}{c}{KL divergence between checkpoints}\\
		\cmidrule{2-3}
		 & 0\% - 10\% & 10\% - 100\% \\
		\midrule
		AL & 0.0691 $\pm$  0.0035 & 0.0451 $\pm$ 0.0013 \\
		AAL & $\mathbf{0.0746 \pm 0.0043}$ & $\mathbf{0.0284 \pm 0.0057}$ \\
		\bottomrule \\
	\end{tabular}
	\caption{The quantitative results measuring robustness against the distribution shift problem for the AL and AAL paradigms. See the text in section \ref{section:app_shift} for more details. These scores are obtained by averaging over 5 runs}
	\label{CIFAR_10_KL}
\end{table}

\subsubsection{On the comparison with supervised learning}

Primarily,  we covered three types of learning paradigms: the supervised learning with standard static train/test splits, the active learning (AL) paradigm and the adaptive active learning (AAL).

In our main result for the KIBA dataset showing by Table~\ref{table2}, the \emph{hybrid} and \emph{greedy} strategies are exclusively devised to be adopted in AL or AAL paradigm.
On the other hand, the \emph{random} strategy can be viewed as equivalent as a supervised learning paradigm built on the standard train-test split.
Being more specific, randomly drawing $n \cdot m$ samples for training is the same as gradually accumulate samples randomly, with $n$ for each batch and $m$ batches overall.

Meanwhile, based on the results in section \ref{section:affinity}, we conclude that both AL- and AAL-based frameworks outperform the supervised learning paradigm by a large margin, with AAL ranking on top.



\subsection{A deeper look at CIFAR-10 experiments}



\begin{figure}[h]
	\centering
	\includegraphics[width=0.75\linewidth]{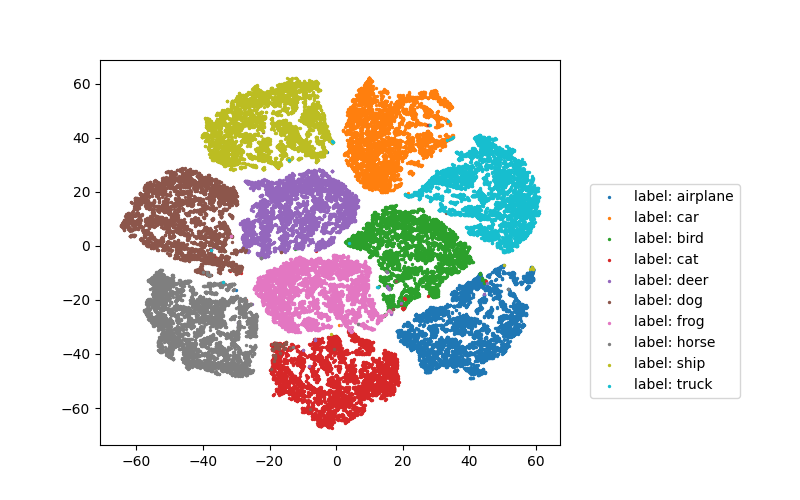}
	\caption{Distribution of CIFAR-10 clusters, visualized with t-SNE applied to the feature vectors attained by a well-trained ResNet.}
	\label{cifar10_cluster}
\end{figure}

\begin{figure}[h]
	\centering
	\includegraphics[width=0.75\linewidth]{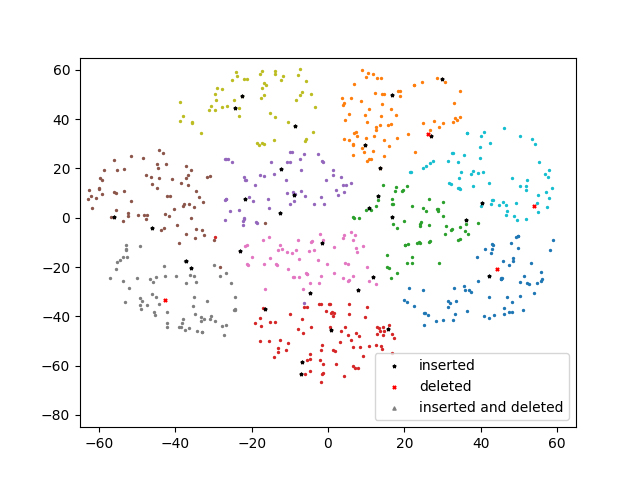}
	\caption{The distribution of the labeled pool, added samples, and deleted samples with AAL at iteration 20. The quantities of samples selected from each class are similar. We also observe that selected samples are closer to decision boundaries, and deleted samples are getting closer to the centroids.}
	\label{cifar10_20}
\end{figure}

\textbf{Addition and deletion.} Figure \ref{cifar10_cluster} shows the distribution of the whole data set whose dimension has been reduced with t-SNE of perplexity 30. In order to perform dimensionality reduction, we use a well-trained ResNet-18\footnote{This well-trained model has reached an accuracy of 99.38\% on the training set, and accuracy of 92.59\% on the test set. The detail of this network has been described in section \ref{resnet18_cifar10_architecture}.} as the feature extractor, and then apply t-SNE to those feature vectors to obtain the 2D distribution of the whole data set. In figure \ref{cifar10_cluster}, we can see samples from 10 different classes are divided into 10 groups\footnote{Since the feature extractor we selected does not reach 100\% accuracy, misclassified data can be observed at different clusters.}. 
Figure \ref{cifar10_20} shows that addition and deletion distributions are different.

\begin{figure}[h]
	\centering
	\includegraphics[width=0.75\linewidth]{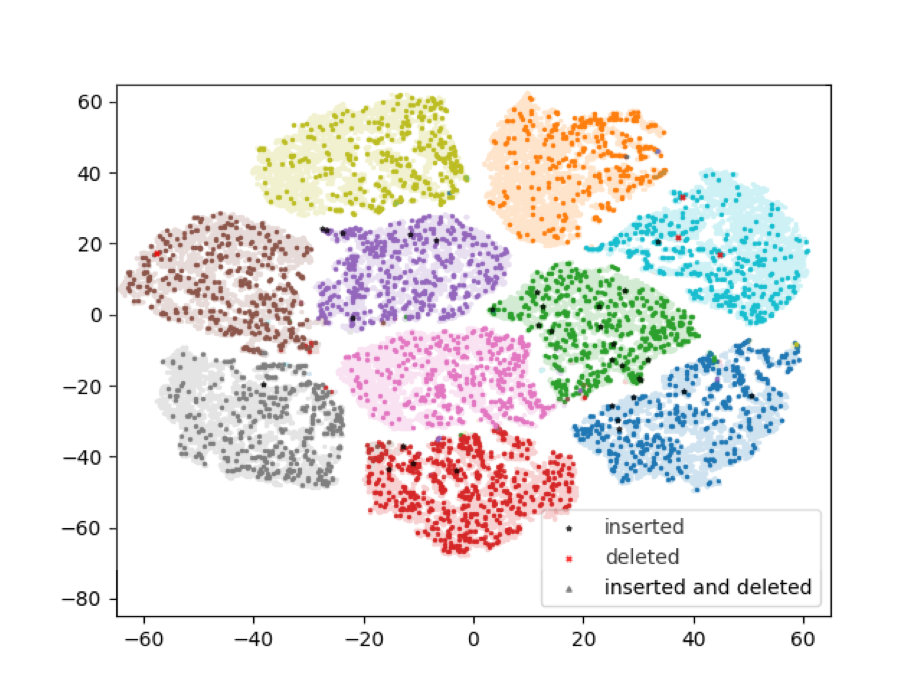}
	\caption{Data kept at last iteration with full data overlay, where the data points with shallower color denote the points in full data, the points with darker color denote the data points in the labeled pool.}
	\label{cifar10_last_full_overlay_add}
\end{figure}

\begin{figure}[h]
	\centering
	\includegraphics[width=0.75\linewidth]{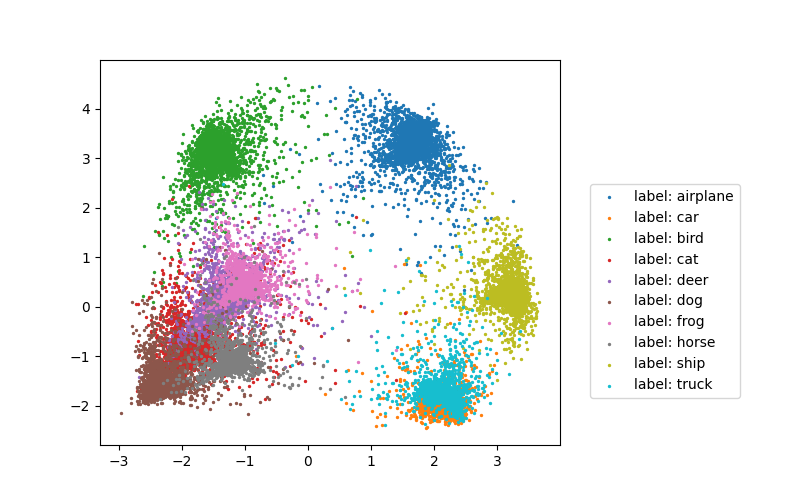}
	\caption{2D PCA on the feature vectors extracted by a well-trained ResNet-18 of each data point in CIFAR-10. Cat(red) data overlap with clusters of dogs, frogs, and horses (brown, pink, and grey), making it a hard target to the learning process.}
	\label{cifar10_class_distributions_pca}
\end{figure}

\begin{figure}[h]
	\centering
	\subfloat[AL, $s=128$]{
		\includegraphics[width=0.18\linewidth]{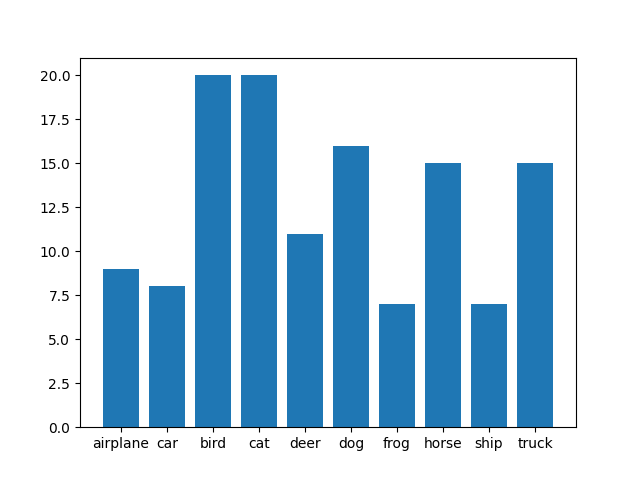}
	}
	\subfloat[AL, $s=268$]{
		\includegraphics[width=0.18\linewidth]{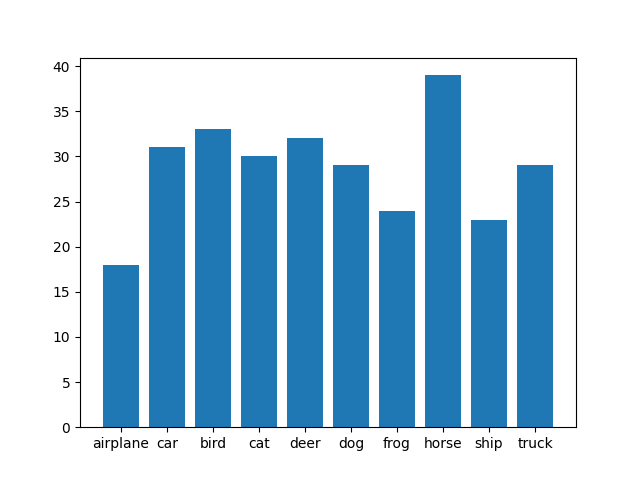}
	}
	\subfloat[AL, $s=1052$]{
		\includegraphics[width=0.18\linewidth]{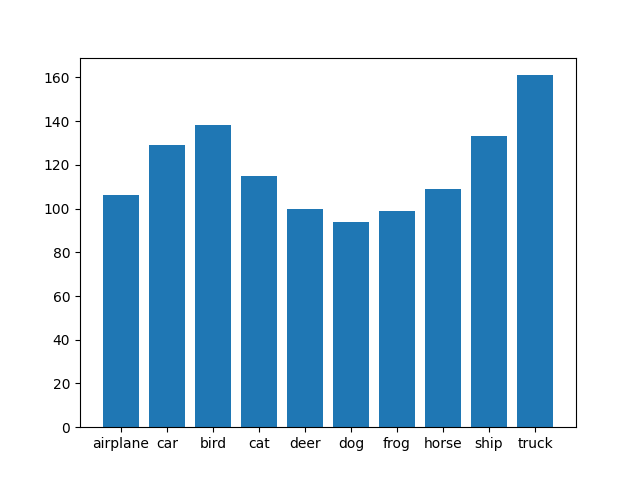}
	}
	\subfloat[AL, $s=4104$]{
		\includegraphics[width=0.18\linewidth]{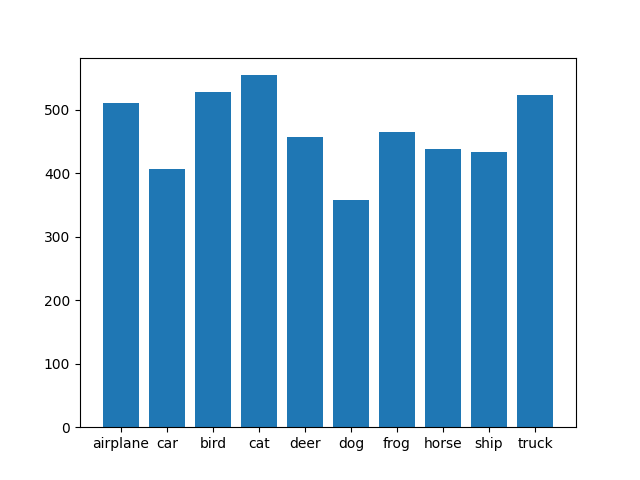}
	}
	\subfloat[AL, $s=8192$]{
		\includegraphics[width=0.18\linewidth]{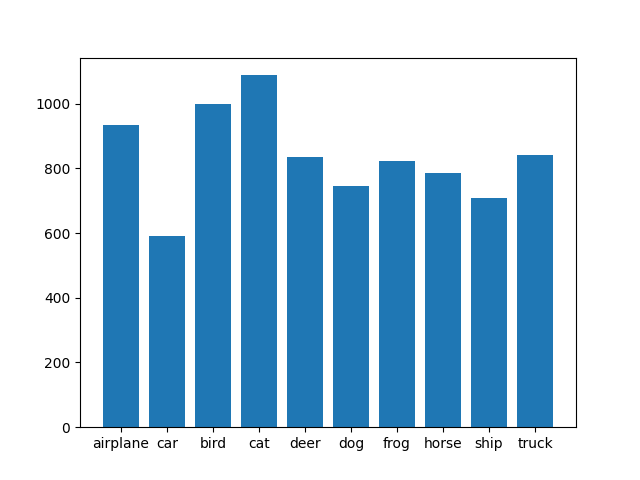}
	} \newline
	\subfloat[AAL, $s=128$]{
		\includegraphics[width=0.18\linewidth]{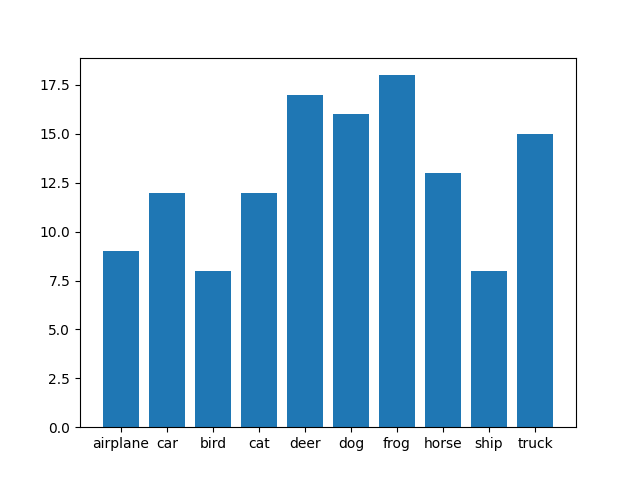}
	}
	\subfloat[AAL, $s=268$]{
		\includegraphics[width=0.18\linewidth]{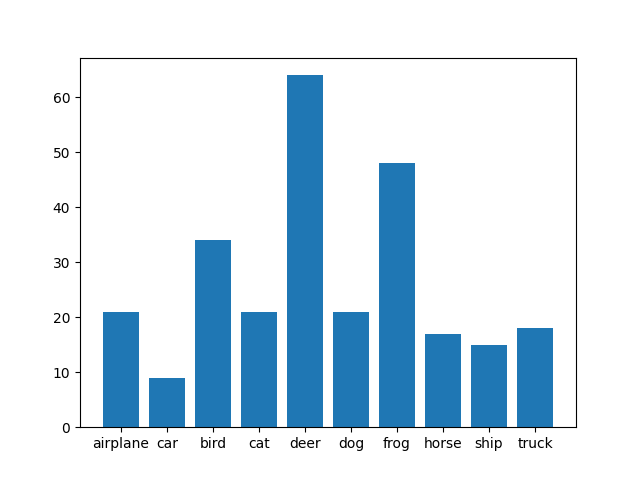}
	}
	\subfloat[AAL, $s=1052$]{
		\includegraphics[width=0.18\linewidth]{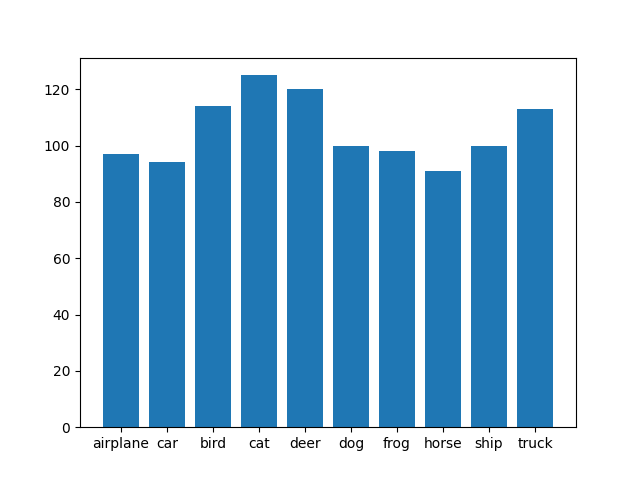}
	}
	\subfloat[AAL, $s=4104$]{
		\includegraphics[width=0.18\linewidth]{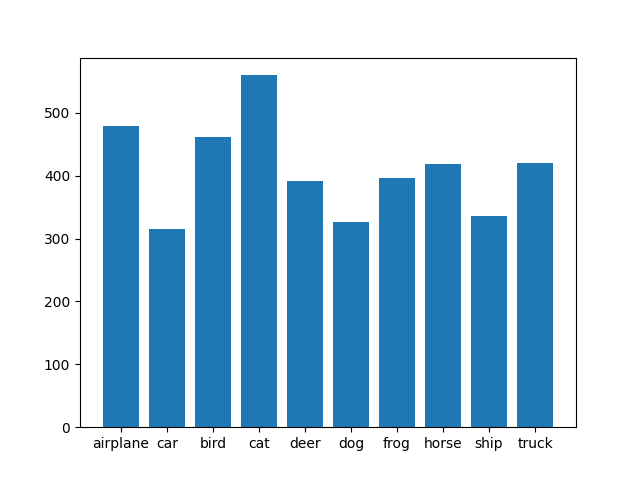}
	}
	\subfloat[AAL, $s=8192$]{
		\includegraphics[width=0.18\linewidth]{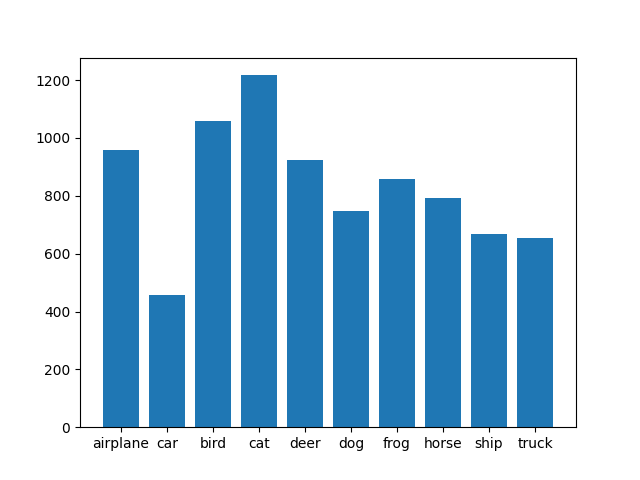}
	}
	\caption{This group of image shows how data distribution changes as AL and AAL iterate, from a randomized initial pool of 128 samples ($s=128$) to 8192 samples.}
	\label{cifar10_class_distribution_changes}
\end{figure}

In CIFAR-10, entropy has been used as the principal criterion for adding samples. We hypothesize that when data points are closer to decision boundaries or more distant from the centroids of clusters, they are more likely to be selected. Figure \ref{cifar10_last_full_overlay_add} shows the distribution of the labeled pool at the final iteration (darker colored) in comparison with the distribution of the full data set (lighter colored). Samples that have been added or deleted in this iteration are also presented. After 20 iterations, we observe a higher density of samples on the decision boundaries than the centroids. We further discuss this pattern in Figure \ref{cifar10_class_distributions_pca} and \ref{cifar10_class_distribution_changes}.

\begin{figure}[h]
	\centering
	\includegraphics[width=0.75\linewidth]{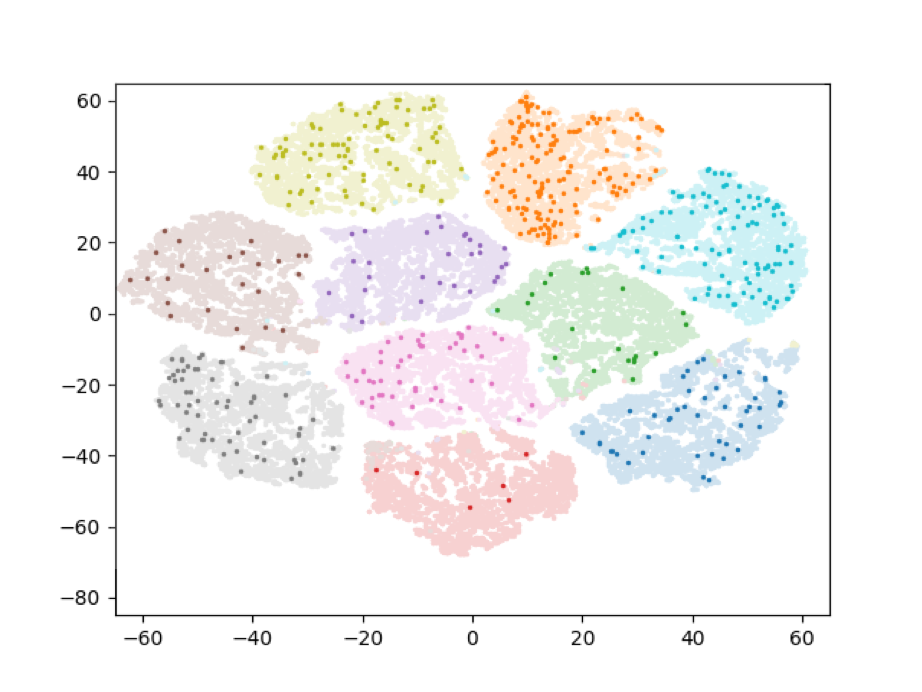}
	\caption{Total data deleted(darker color) until the last iteration overlaid on full data.(ligher color) Notice that the distribution of deleted samples tends to be more uniform and closer to centroids compare to the total added samples on Figure \ref{cifar10_last_full_overlay_add}. }
	\label{Exploration trajectory for a single experiment}
\end{figure}

We chose a combination of entropy and cosine distance to be the strategy for deletion. Under this strategy, we expect the overlapped samples and samples that are closer to the centroids of clusters to be removed. In another word, the most "common" or "typical" samples in the labeled pool should be removed. Figure \ref{Exploration trajectory for a single experiment} shows the distributions of all the removed samples.  We find that the distribution difference between added and deleted samples meet the expectation that deleted data are more uniform and closer to the centroids than added samples.

Figure \ref{cifar10_class_distribution_changes} shows class distribution at the last iteration. At the initial stage, we randomly choose 128 samples, after which we add 32 samples per iteration in both AL and AAL, and delete 4 samples per iteration in AAL. We observed two patterns:

\begin{enumerate}
    \item Both AL and AAL tend to select more samples with label \texttt{cat}. According to figure \ref{cifar10_class_distributions_pca}, a lot of cat data overlapped with clusters of dogs, frogs, and horses, which makes cats much harder to learn. In Figure \ref{Exploration trajectory for a single experiment} we also find that very few cat samples are removed.
    \item Compared to AL, AAL can have a more drastic distribution shift at the first few iterations but converges faster to a fixed distribution at the last iterations. This finding is also further confirmed by the KL divergence tests from Table \ref{CIFAR_10_KL}.
\end{enumerate} 


\subsection{Data and training details}

\subsubsection{KIBA}
 \textbf{Dataset.} The KIBA data pipeline is adopted from a previous study \cite{ozturk2018deepdta}. KIBA includes 229 proteins and 2111 drugs and 118254 KIBA (interaction) score between proteins and drugs. We represent each protein and drug as a one-hot vector. 
 
\textbf{Model structure. } The model includes two encoders for both drug and protein. The drug one-hot vector is encoded to a 128-dimensional vector with a three-layer feedforward network and the protein vector is transformed to a target dimension of 128 with a linear layer. The affinity score is calculated as the dot product of drug and protein vectors. The model is trained with a batch size of 64, and an SGD optimizer with a 0.001 learning rate during each iteration.

\textbf{Training.} For each AL and AAL iteration, the model is trained from scratch with the currently selected batch of data. To prevent overfitting, we used an early stopping rule. The training will stop if the validation accuracy does not improve within 3 epochs. For uncertainty calculation, we trained 5 different models in each iteration to form a model committee. Then the trained model is used to calculate corresponding statistics used for selecting the next batch of data. In the first iteration, 64 samples are randomly selected as a starting batch. Then every iteration, AL will pick another 64 data which ranked highest among the designated metrics. For AAL, the 8 worst ranking data within the existing training data will be discarded. The selection and training processes constitute an iteration. The iteration stops when the coverage score reaches 0.3 or the total number of iterations reaches 300. A complete 300 iteration run takes around 3 hours on an NVIDIA RTX 2080Ti GPU.

\subsubsection{CIFAR-10}

\textbf{Dataset.} The CIFAR-10 dataset consists of 50,000 training and 10,000 testing examples, all of which are 32x32 RGB images drawn from 10 classes. We use 5,000 training examples as validation examples.

\textbf{Training.} For experiments conducted on CIFAR-10 with ResNet-18, we optimize parameters using SGD with an initial learning rate of 0.1, with momentum 0.9, and a weight decay of 0.0001. For each cycle, we train our models with mini-batches of size 64. Before conducting an experiment, an upper bound(4096, 8192) of the labeled pool was given. We initialize the labeled pool with 128 randomly selected samples. For each cycle, we add 32 samples to the labeled pool in AL, add 32 samples to the labeled pool and delete 4 samples from the labeled pool in AAL, and train the models with this labeled pool for 10 epochs, and reuse the parameters obtained from the last cycle. If there is no increase in performance in 5 epochs inside a cycle, this cycle terminates and the parameters acquired in this cycle would be represented by the parameters in the last cycle. The training iteration continues until the given upper bound is reached. Random horizontal flip with a probability of 0.5 and normalization are applied to CIFAR-10. No pre-training is performed in our experiments. When applying AL, we use entropy as the only uncertainty score for addition. When applying AAL, we use entropy for addition. For deletion, we use a ratio of 1:1 combination of entropy and distance, in which entropy is defined as Shannon Entropy, and distance is defined as the euclidean distances from each point to the nearest centroid obtained by KMeans. To add more randomness, when selecting top uncertain samples(or deleting top certain samples), we randomly choose $n$ samples in top $2n$ samples to add(or delete).

\textbf{Feature extractor} \label{resnet18_cifar10_architecture} For CIFAR-10, we used a modified ResNet-18. Since CIFAR-10 images are too small, down-sampling is not suitable in training CIFAR-10 images. Thus, we have removed the down-sampling in "\_make\_layer", and also the classifier component from the official PyTorch implementation. The result is a ResNet-18 feature extractor. 

\subsection{Addtional Figures}

\begin{figure}[h]
	\centering
	\subfloat{
		\includegraphics[width=0.35\linewidth]{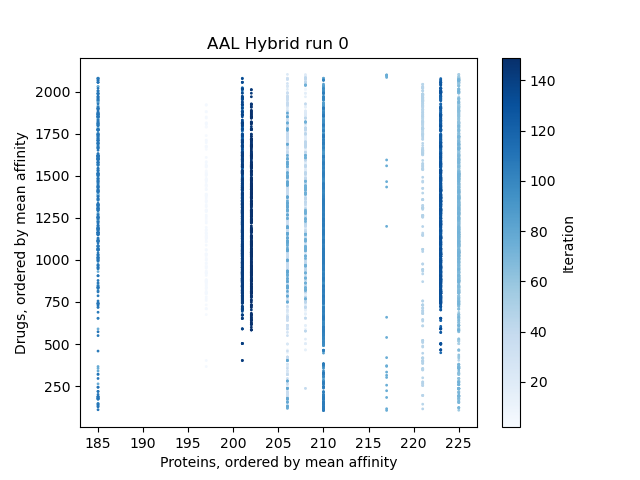}
	} \centerhfill
	\subfloat{
		\includegraphics[width=0.35\linewidth]{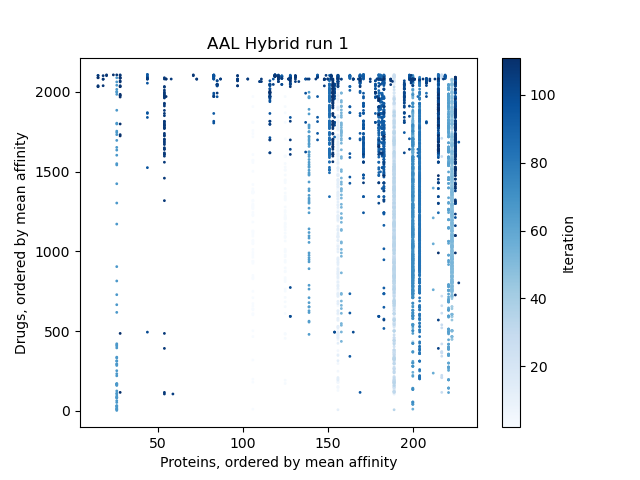}
	} \newline
	\subfloat{
		\includegraphics[width=0.35\linewidth]{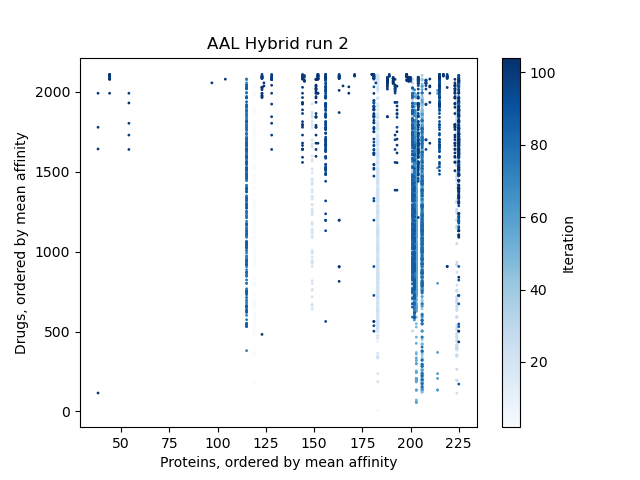}
	} \centerhfill
	\subfloat{
		\includegraphics[width=0.35\linewidth]{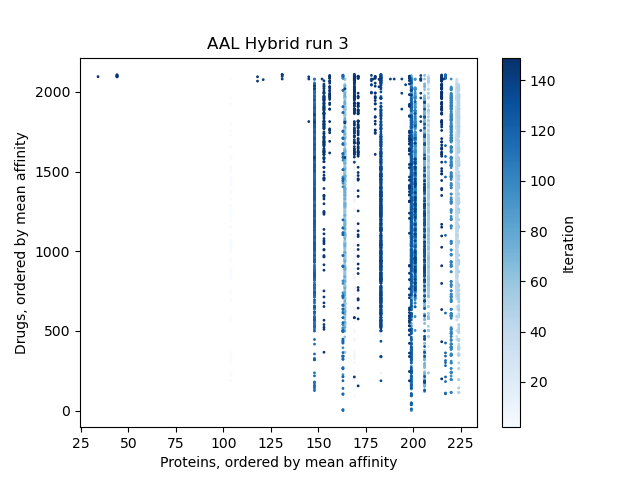}
	} \newline
	\subfloat{
		\includegraphics[width=0.35\linewidth]{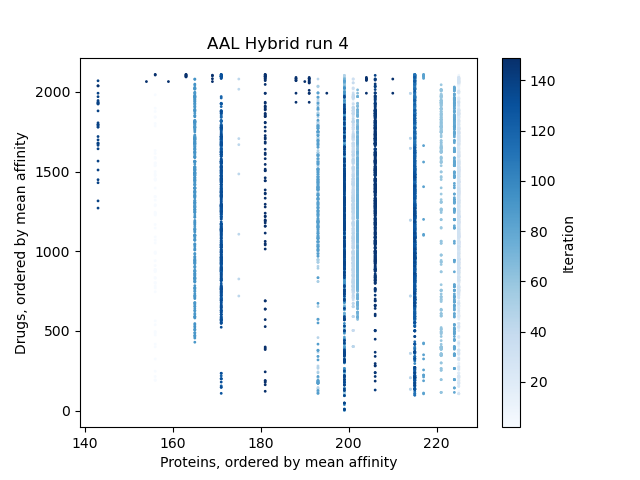}
	} \centerhfill
	\subfloat{
		\includegraphics[width=0.35\linewidth]{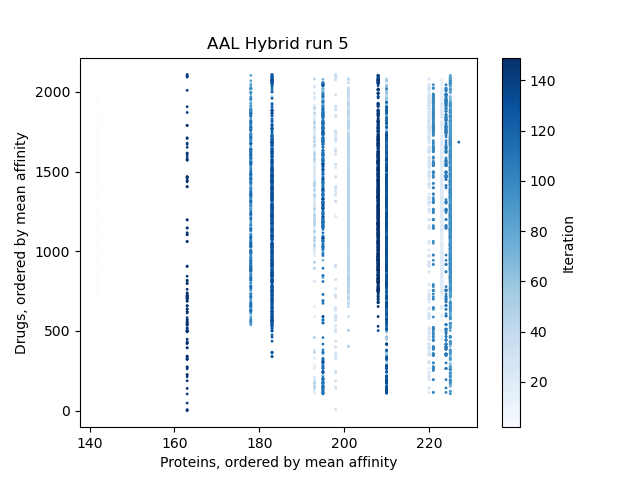}
	} \newline
	\subfloat{
		\includegraphics[width=0.35\linewidth]{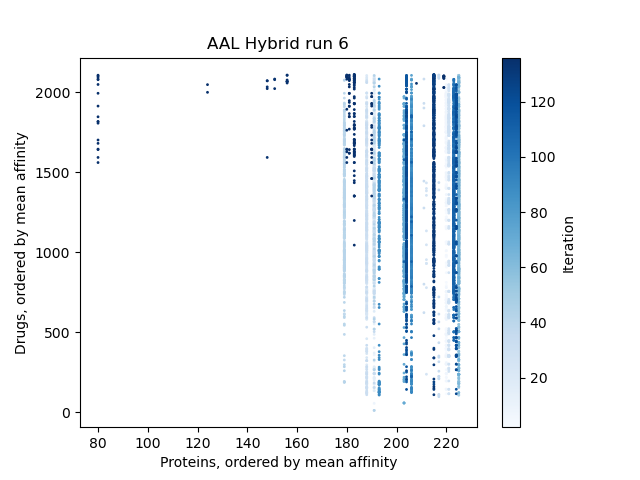}
	} \centerhfill
	\subfloat{
		\includegraphics[width=0.35\linewidth]{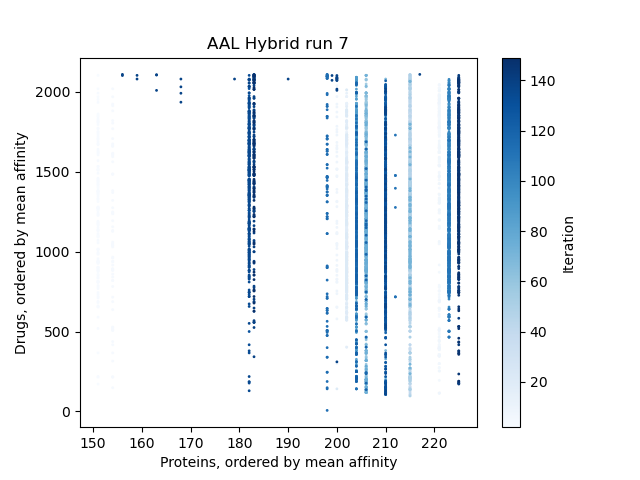}
	} \newline
	\subfloat{
		\includegraphics[width=0.35\linewidth]{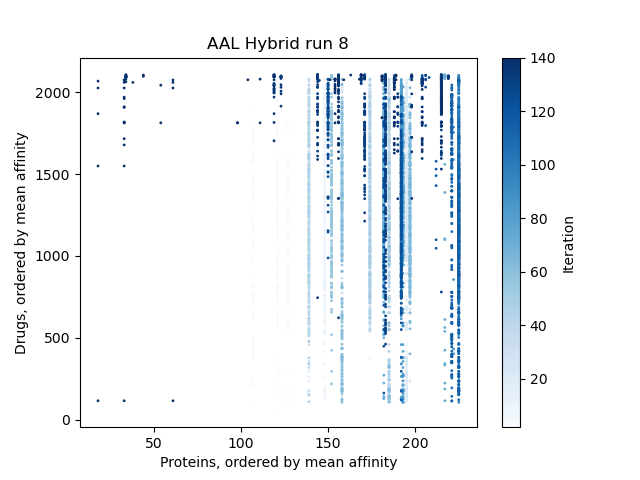}
	} \centerhfill
	\subfloat{
		\includegraphics[width=0.35\linewidth]{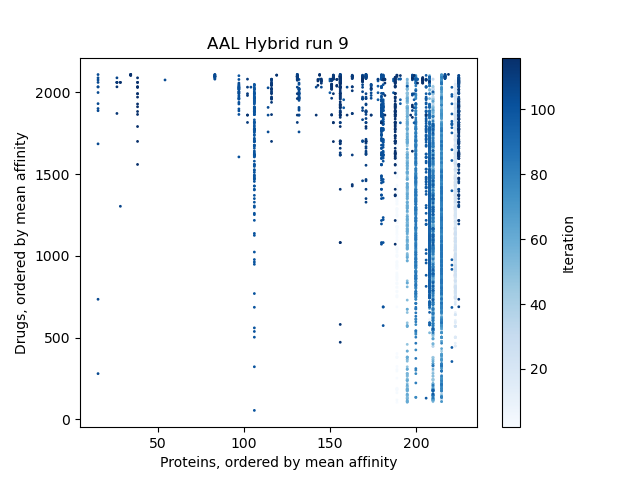}
	} \newline
	\caption{Progression for all 10 runs with AAL Hybrid}
	\label{aal_hybrid}
\end{figure}

\begin{figure}[h]
	\centering
	\subfloat{
		\includegraphics[width=0.35\linewidth]{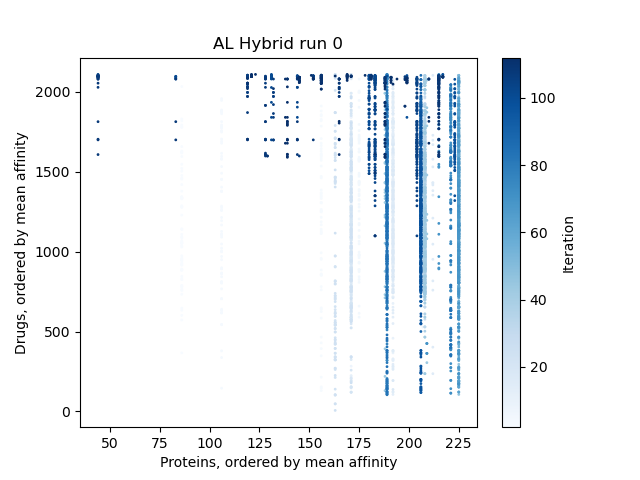}
	} \centerhfill
	\subfloat{
		\includegraphics[width=0.35\linewidth]{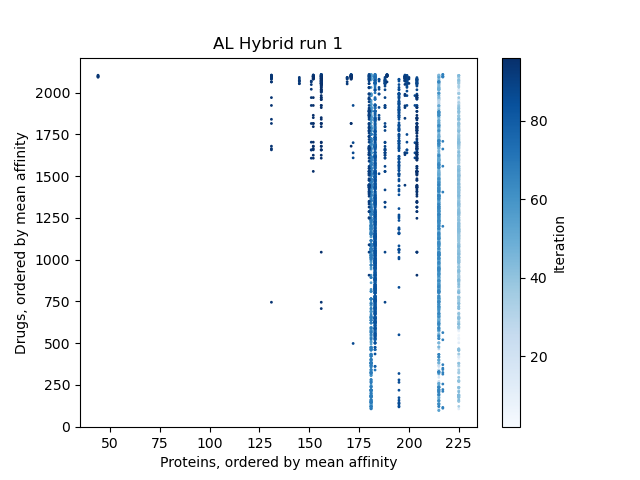}
	} \newline
	\subfloat{
		\includegraphics[width=0.35\linewidth]{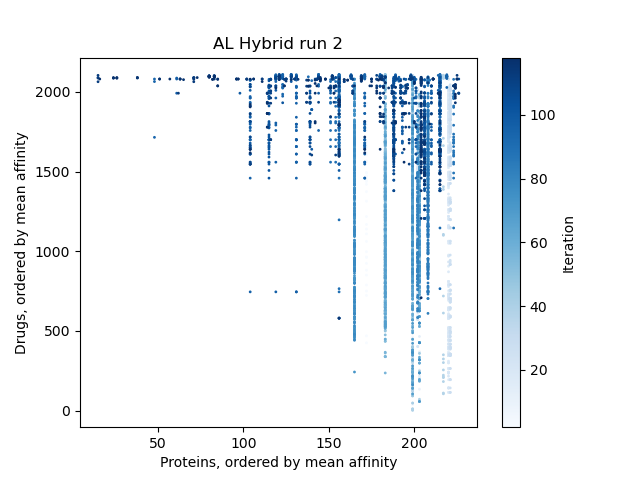}
	} \centerhfill
	\subfloat{
		\includegraphics[width=0.35\linewidth]{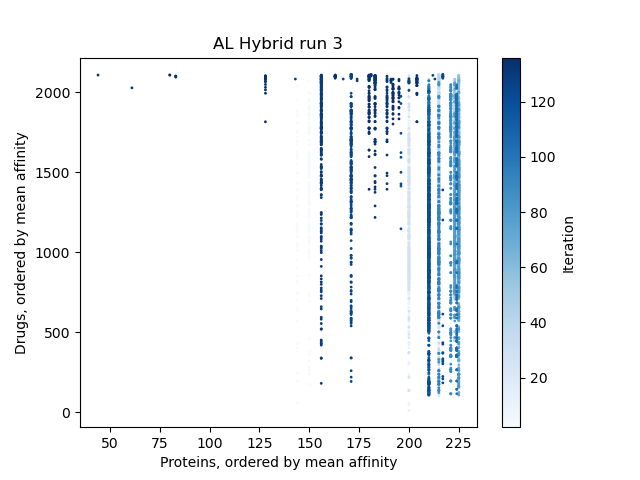}
	} \newline
	\subfloat{
		\includegraphics[width=0.35\linewidth]{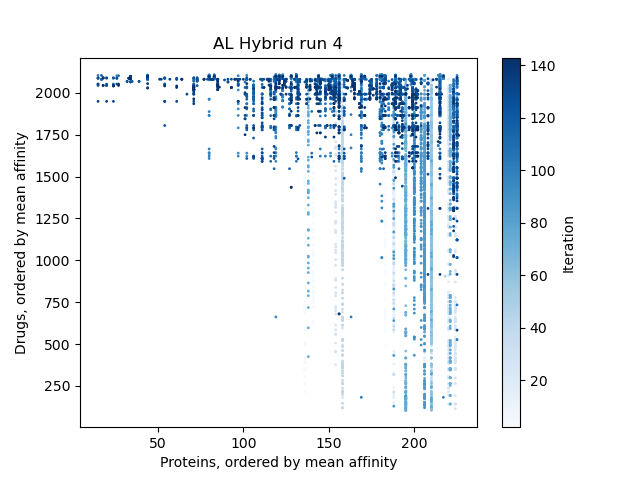}
	} \centerhfill
	\subfloat{
		\includegraphics[width=0.35\linewidth]{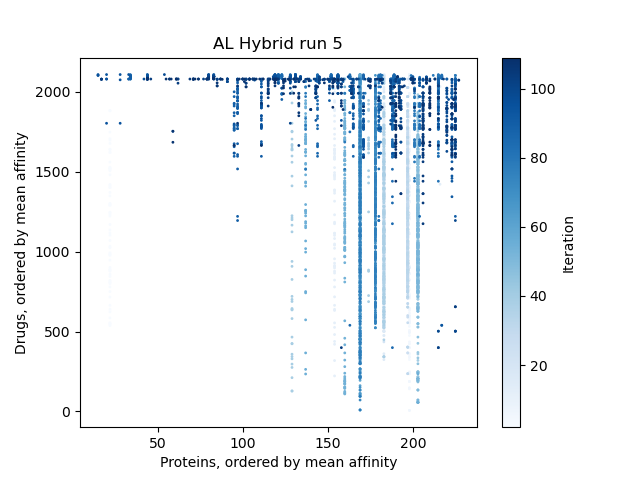}
	} \newline
	\subfloat{
		\includegraphics[width=0.35\linewidth]{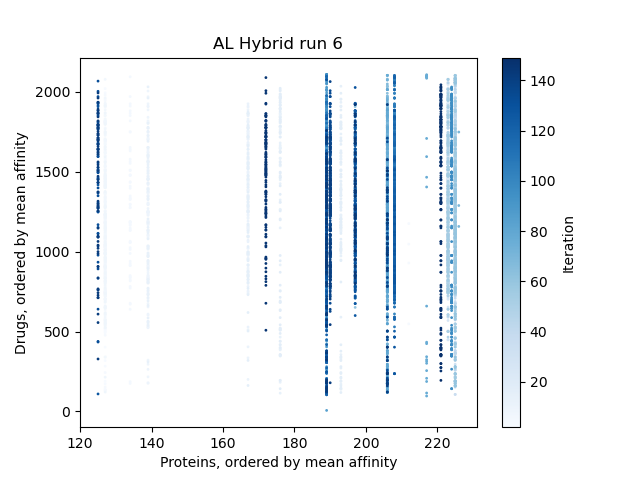}
	} \centerhfill
	\subfloat{
		\includegraphics[width=0.35\linewidth]{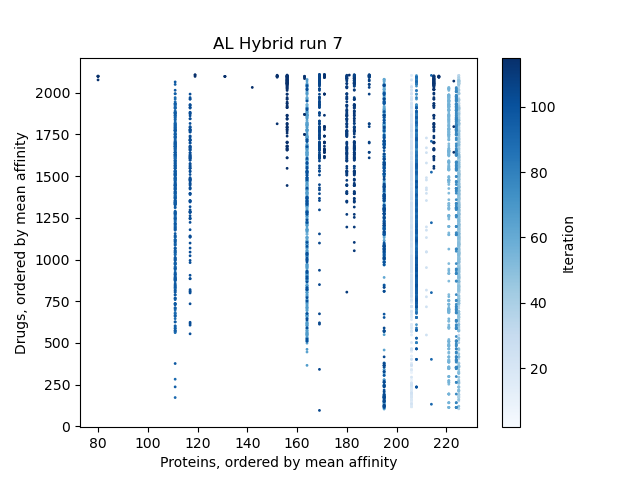}
	} \newline
	\subfloat{
		\includegraphics[width=0.35\linewidth]{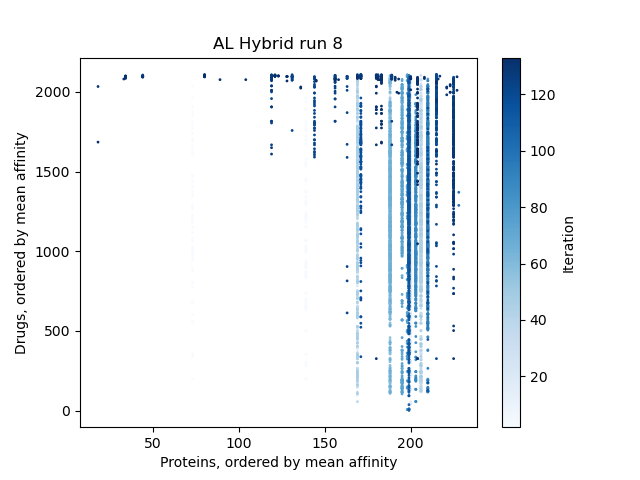}
	} \centerhfill
	\subfloat{
		\includegraphics[width=0.35\linewidth]{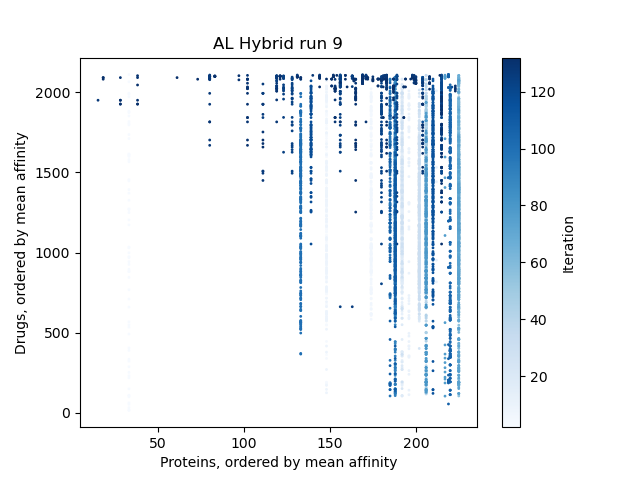}
	} \newline
	\caption{Progression for all 10 runs with AL Hybrid}
	\label{al_hybrid}
\end{figure}

\begin{figure}[h]
	\centering
	\subfloat{
		\includegraphics[width=0.35\linewidth]{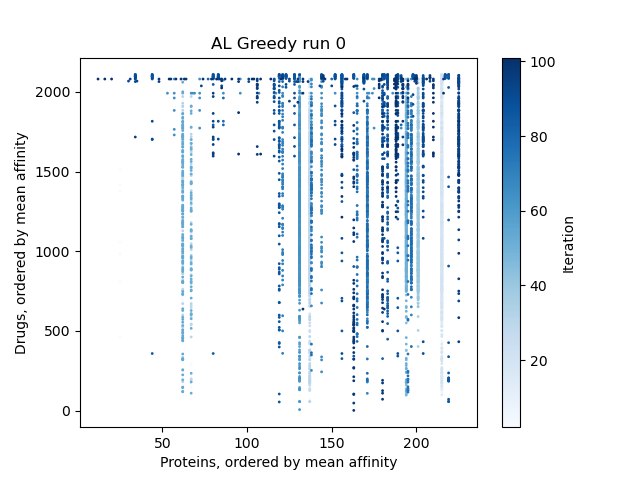}
	} \centerhfill
	\subfloat{
		\includegraphics[width=0.35\linewidth]{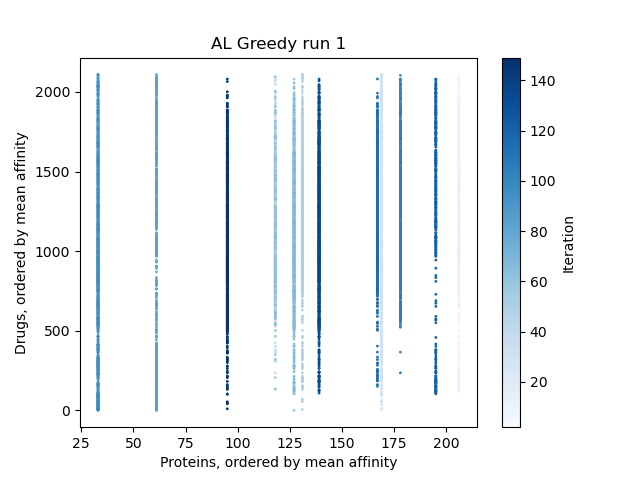}
	} \newline
	\subfloat{
		\includegraphics[width=0.35\linewidth]{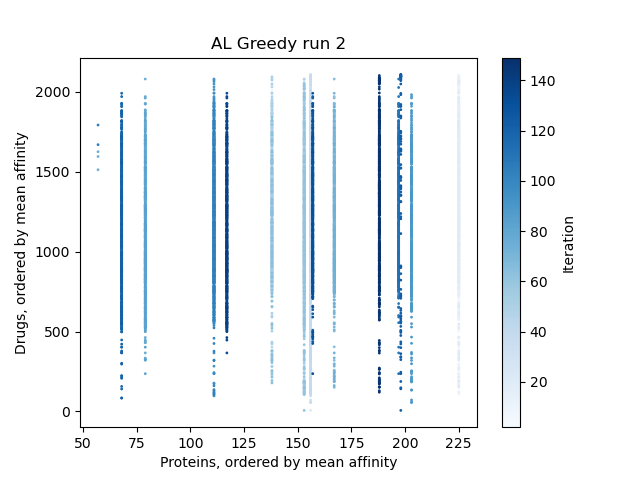}
	} \centerhfill
	\subfloat{
		\includegraphics[width=0.35\linewidth]{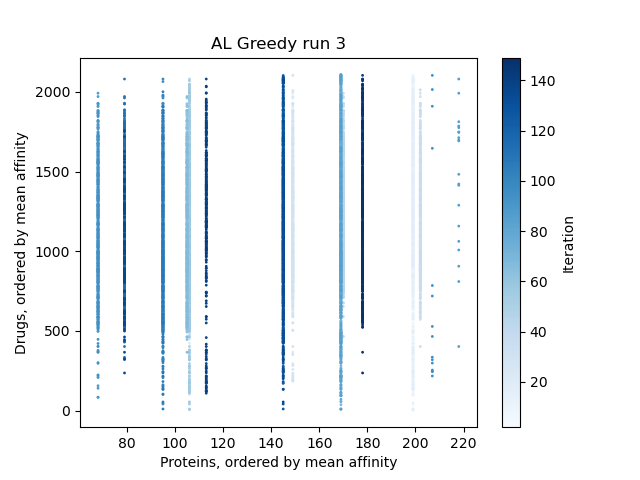}
	} \newline
	\subfloat{
		\includegraphics[width=0.35\linewidth]{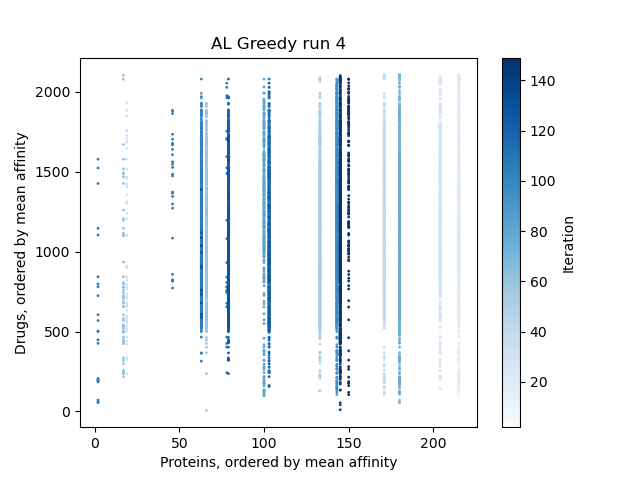}
	} \centerhfill
	\subfloat{
		\includegraphics[width=0.35\linewidth]{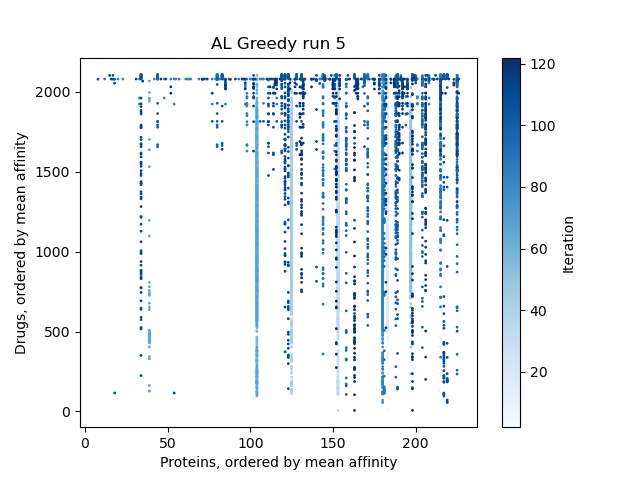}
	} \newline
	\subfloat{
		\includegraphics[width=0.35\linewidth]{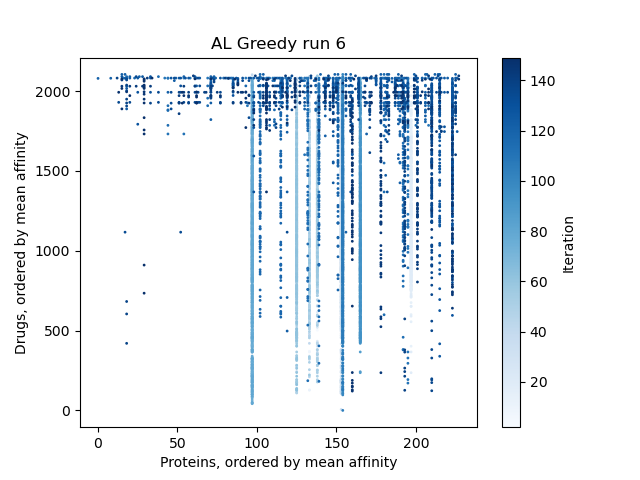}
	} \centerhfill
	\subfloat{
		\includegraphics[width=0.35\linewidth]{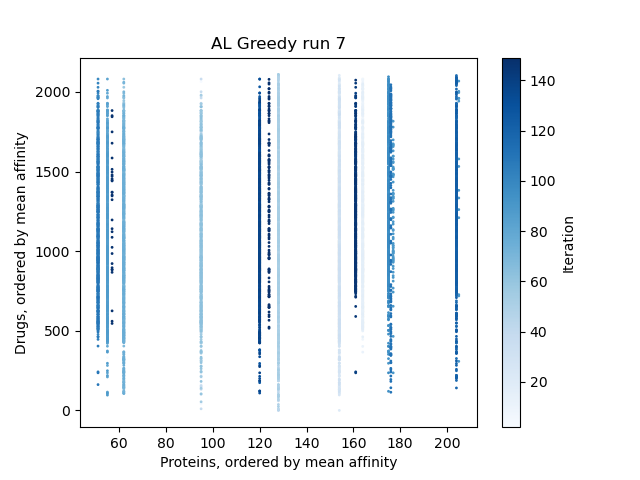}
	} \newline
	\subfloat{
		\includegraphics[width=0.35\linewidth]{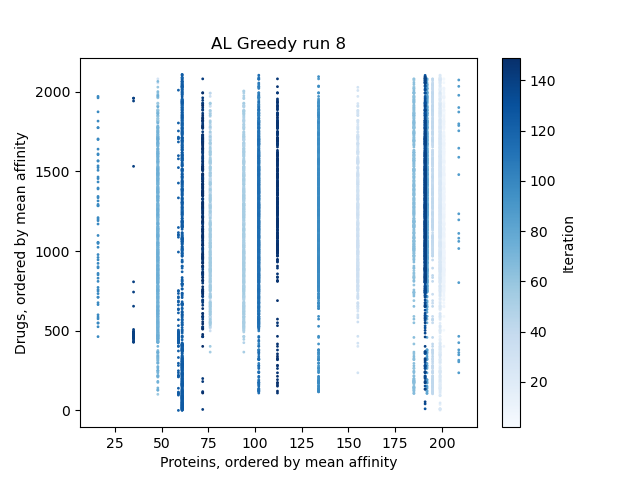}
	} \centerhfill
	\subfloat{
		\includegraphics[width=0.35\linewidth]{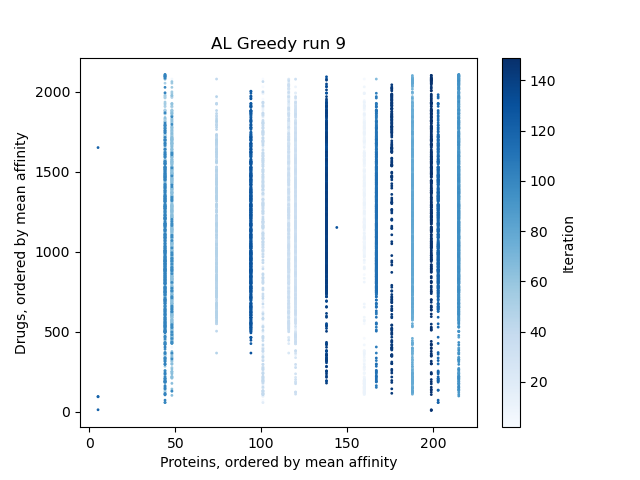}
	} \newline
	\caption{Progression for all 10 runs with AL Greedy}
	\label{al_greedy}
\end{figure}

\begin{figure}[h]
	\centering
	\subfloat{
		\includegraphics[width=0.35\linewidth]{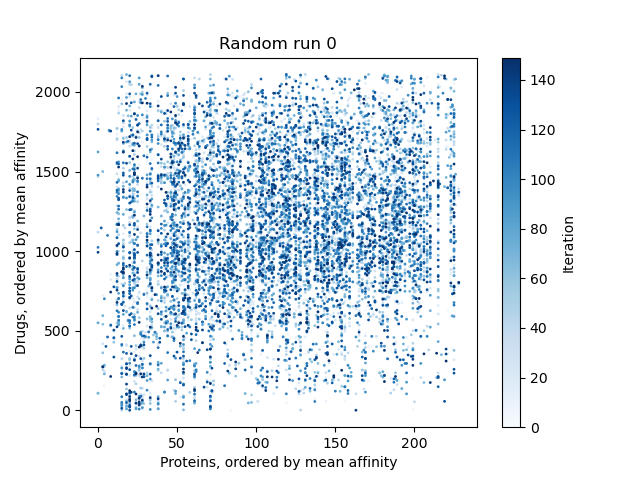}
	} \centerhfill
	\subfloat{
		\includegraphics[width=0.35\linewidth]{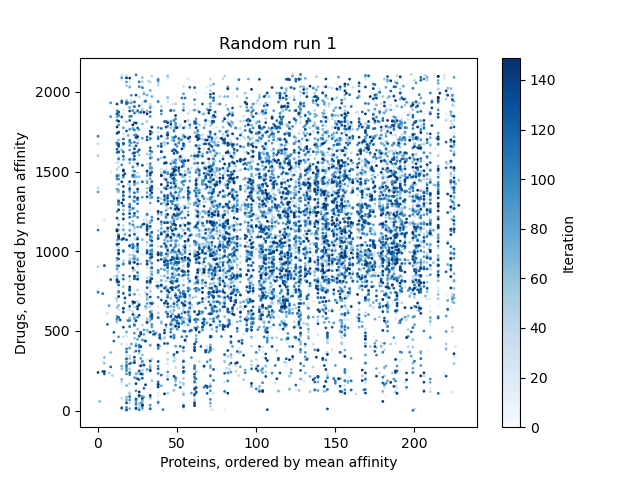}
	} \newline
	\subfloat{
		\includegraphics[width=0.35\linewidth]{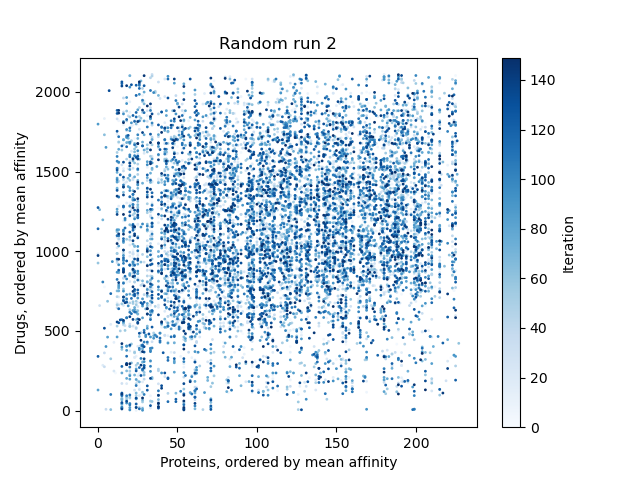}
	} \centerhfill
	\subfloat{
		\includegraphics[width=0.35\linewidth]{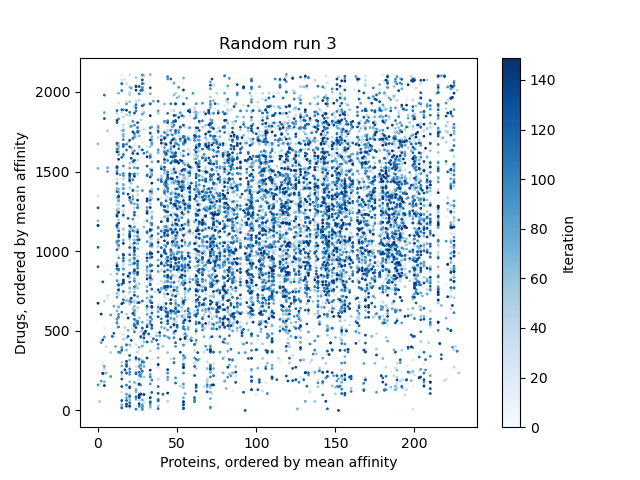}
	} \newline
	\subfloat{
		\includegraphics[width=0.35\linewidth]{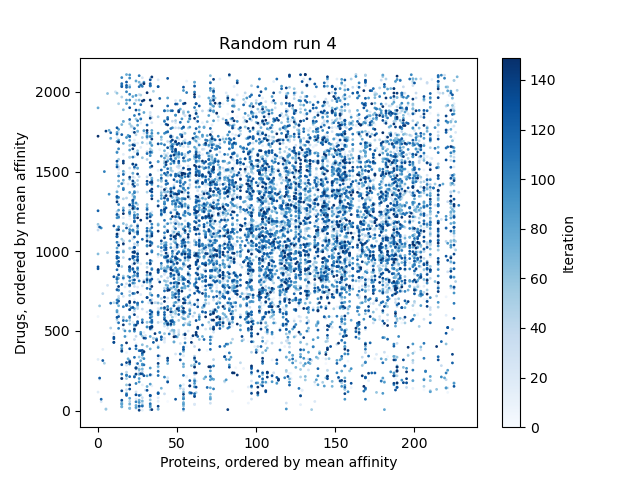}
	} \centerhfill
	\subfloat{
		\includegraphics[width=0.35\linewidth]{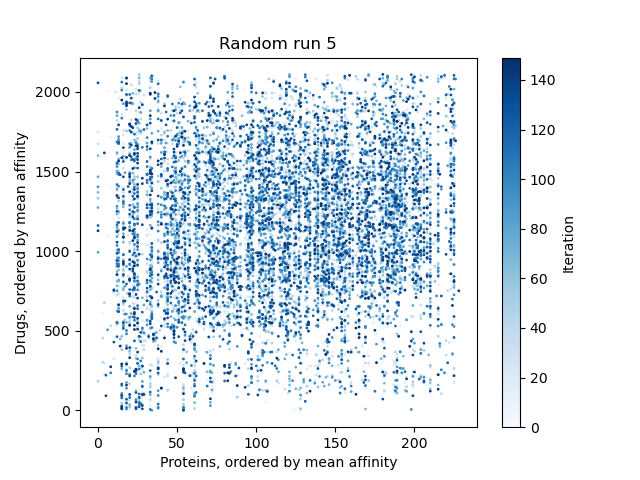}
	} \newline
	\subfloat{
		\includegraphics[width=0.35\linewidth]{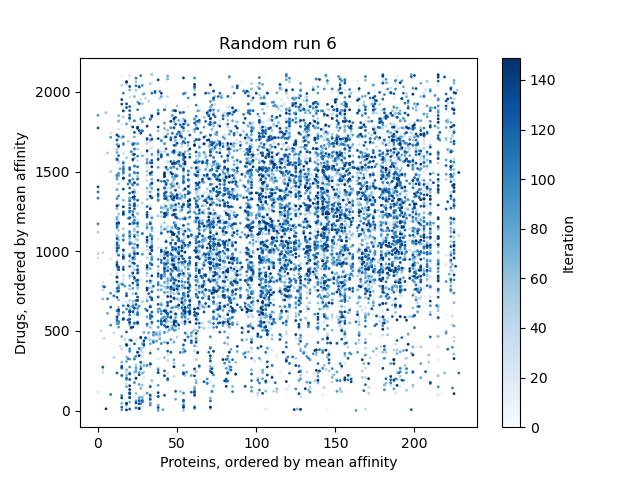}
	} \centerhfill
	\subfloat{
		\includegraphics[width=0.35\linewidth]{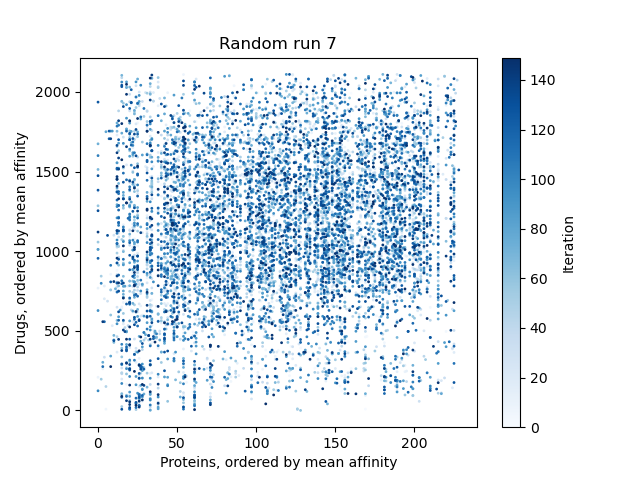}
	} \newline
	\subfloat{
		\includegraphics[width=0.35\linewidth]{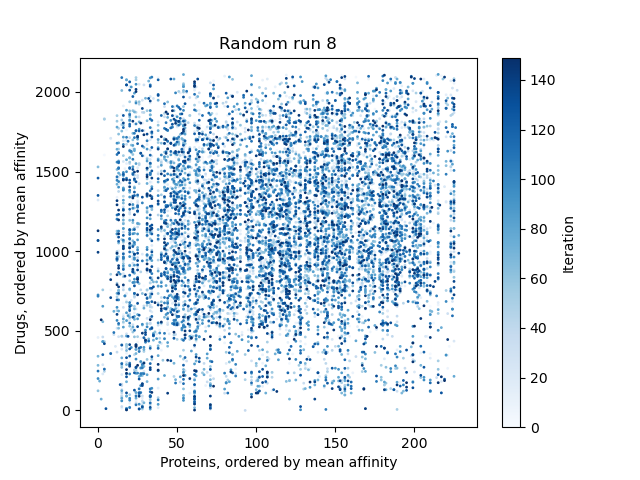}
	} \centerhfill
	\subfloat{
		\includegraphics[width=0.35\linewidth]{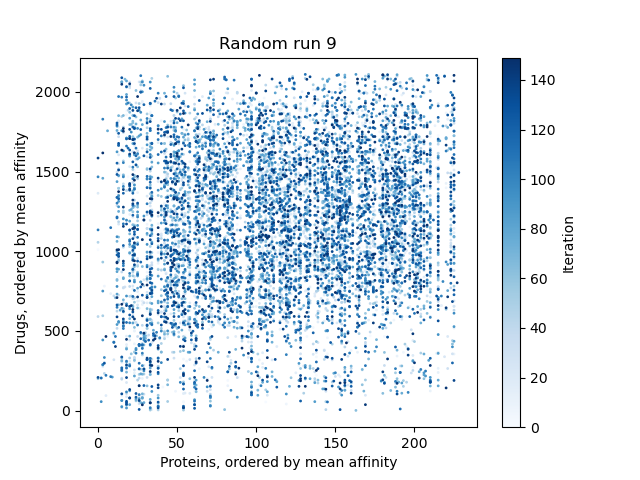}
	} \newline
	\caption{Progression for all 10 runs with AL Random}
	\label{al_random}
\end{figure}